\documentclass[journal]{IEEEtran}
\ifCLASSINFOpdf
\else
\fi
\usepackage{graphicx}
\usepackage{epsfig}
\usepackage{epstopdf}
\usepackage{float}
\usepackage{fancyhdr}



\begin{document}
%
\title{The Labeled Multiple Canonical Correlation Analysis for Information Fusion}
%
%
%

\author{Lei~Gao,~
        Rui~Zhang,~
        Lin~Qi,~
        Enqing~Chen,~\IEEEmembership{Member,~IEEE,}
        and~Ling~Guan,~\IEEEmembership{Fellow,~IEEE,}
\thanks{L. Gao is with the Department of Electrical and Computer Engineering, Ryerson University, Toronto, ON, M5B 2K3, Canada (iegaolei@gmail.com).}
\thanks{R. Zhang is currently a senior researcher at the Epson Canada, Toronto, Canada (ray.rui.zhang@gmail.com).}
\thanks{L. Qi is with the school of Information Engineering, Zhengzhou University, Zhengzhou, China (ielqi@zzu.edu.cn).}
\thanks{E. Chen is with the school of Information Engineering, Zhengzhou University, Zhengzhou, China (ieeqchen@zzu.edu.cn).}
\thanks{L. Guan is with the Department of Electrical and Computer Engineering, Ryerson University, Toronto, ON, M5B 2K3, Canada (lguan@ee.ryerson.ca).}}
\maketitle

\begin{abstract}
The objective of multimodal information fusion is to mathematically analyze information carried in different sources and create a new representation which will be more effectively utilized in pattern recognition and other multimedia information processing tasks. In this paper, we introduce a new method for multimodal information fusion and representation based on the Labeled Multiple Canonical Correlation Analysis (LMCCA). By incorporating class label information of the training samples, the proposed LMCCA ensures that the fused features carry discriminative characteristics of the multimodal information representations, and are capable of providing superior recognition performance. We implement a prototype of LMCCA to demonstrate its effectiveness on handwritten digit recognition, face recognition and object recognition utilizing multiple features, bimodal human emotion recognition involving information from both audio and visual domains. The generic nature of LMCCA allows it to take as input features extracted by any means, including those by deep learning (DL) methods. Experimental results show that the proposed method enhanced the performance of both statistical machine learning (SML) methods, and methods based on DL.

\end{abstract}


%
\IEEEpeerreviewmaketitle

\section{Introduction}
Since multimodal data contains rich and complimentary information about the semantics presented in the sensory and media data, effective interpretation and integration of multimodal information are quite central for the efficacious utilization of multimedia in a wide variety of applications, such as multiple cameras based event summarization [68] etc. Thus, information fusion is becoming an increasingly important topic in multimedia research communities [1]. In general, there are three categories of information fusion, applied at different levels of the information flow, i.e.  feature/data level, score level and decision level [2]. Feature/data level fusion combines the original data or extracted features through certain strategies before classification [3]. Fusion at the score level combines the outputs of multiple intermediate classifiers or mid-level feature extractions, each dealing with a single modality, through a rule based scheme or by taking them as new features of a different classification algorithm [4]. The decision level fusion usually generates the final results by aggregating the outputs of the classifiers of different modalities using rule based methods [5].\\\indent Since feature level contains richer information about the raw data, it is expected to perform better in comparison with fusion at score level and decision level [6]. Therefore, feature level fusion has drawn more attention from the multimedia community recently. Although rapid and impressive progress has been made in feature level fusion, the identification of the discriminative representation among various modalities, and the design of a fusion strategy that effectively utilizes the complementary information remain a challenge. A wide variety of methods have been proposed to address this challenge [7-8]. Arguably, the first attempt is serial feature fusion [9-10] which directly groups multiple features into one union-vector. Though it is simple, it brings out very limited discriminative information embedded in the extended feature space for better recognition. Since then feature fusion has rapidly evolved into a science based on statistics, in particular the theory of correlation.\\\indent The objective of correlation analysis is to identify and measure the intrinsic association across different modalities, by which the useful information carried by all modalities pertaining to certain semantics can be determined. To this end, canonical correlation analysis (CCA) becomes a natural choice. It has been applied to direct blind channel equalization, computer vision, neural networks and speech recognition [11-13]. In [14-16], generalized canonical correlation analysis, a.k.a. GCCA is proposed to deal with information fusion analysis. Since the supervised information (intra-class information) is introduced to GCCA, it improves the final recognition performance.\\\indent However, since CCA and GCCA only deal with the mutual relationship between two random vectors, limiting the application of these techniques. Thus, as a natural extension of CCA, multiple canonical correlation analysis (MCCA) [17] was proposed to deal with the fusion of more than two modalities/features. It has been used in joint blind source separation [18-19], the study of functional magnetic resonance imaging (fMRI) datasets [20] and so forth. Nevertheless, MCCA does not explore the discriminatory representation and is not capable of providing satisfactory recognition performance. Then, a discriminative multiple canonical correlation analysis (DMCCA) method [58] was presented to extract discriminatory representations from the input data sources, providing an elegant framework for information fusion based on CCA principles.\\\indent In recent years, profiting from the dramatically increased chip processing abilities (e.g. GPU units) and the significantly lowered cost of computing hardware, deep learning (DL) based methods have dramatically improved the state-of-the-art in visual object recognition, object detection and many other domains. Various DL approaches have been extensively reviewed and discussed [42, 43, 49, 50]. Among these methods, Convolutional Neural Networks (CNN) is one of the most notable DL approaches. Hinton et al. [42] proposed a general CNN architecture for image classification. It employed two distinct forms of data augmentation: the first form of data augmentation consists of generating image translations and horizontal reflections, and the second form consists of altering the intensities of the RGB channels in training images. Chan et al. [43] proposed a new DL architecture named principal components analysis network (PCANet). Experimental results demonstrate the potential of PCANet to serve as a simple but highly competitive baseline for texture classification and object recognition.\\\indent In this paper, we propose a generic information fusion method, labeled multiple canonical correlation analysis (LMCCA), to improve the final performance, which makes use of the intra-class scatter matrix of training samples and the cross-correlation matrix of multiple variables to extract the discriminant information. The contributions are summarized as follows.\\
\textbf{1}. We analytically discovered and experimentally verified the property on the relationship between the upper limit on the number of projected dimensions of LMCCA and the best recognition performance it can achieve. The optimally projected dimension in LMCCA has no relation to the number of training samples. \\
\textbf{2}. We verify that LMCCA is the generalized form of canonical correlation analysis (CCA), multiple canonical correlation analysis (MCCA) and the recently proposed generalized canonical correlation analysis (GCCA), hence establishing a unified framework for canonical correlation analysis with intra-class information.\\
\textbf{3}. Due to the generic nature of LMCCA, it is capable of taking as input features extracted by any means, classical features or those by DL methods. Such methodological fusion evidently improves recognition performance of these features. The effectiveness and generality of LMCCA are experimentally demonstrated by performance evaluation on several application examples.\\\indent The remainder of this paper is organized as follows. The theory of LMCCA is derived in Section II. In Section III, feature extraction and implementation of LMCCA for different applications are presented. The experimental results and analysis are given in Section IV. Conclusions are drawn in Section V.
\section{The Theory of LMCCA}
In what follows, we first briefly present the definitions of CCA, GCCA $\&$ MCCA, and then move on to the formulation of LMCCA.
\subsection{Fundamentals of CCA}
The objective of CCA is to find basis vectors for two sets of variables such that the correlation between the projections of the variables onto these basis vectors are mutually maximized. Therefore, all useful information is maximumly preserved through the projections.\\\indent Let $ x_1 \in {R^{\;m}},x_2 \in {R^{\;p }} $ be two sets of variables as the entries. CCA finds a pair of directions $ \omega_{x_1} $ and $ \omega_{x_2} $ that maximize the correlation between the projections of the two canonical vectors: $ X_1 = \omega_{x_1}^Tx_1  $, $ X_2 = \omega_{x_2}^Tx_2  $, which is mathematically expressed as:
\begin{equation} \mathop {\arg \max }\limits_{{\omega _{x_1}},{\omega _{x_2}}} {\omega _{x_1}}^T{R_{x_1x_2}}{\omega _{x_2}},\end{equation} where ${R_{x_1x_2}} = x_1{x_2^T}$ is the cross-correlation matrix of the vectors $ x_1 $, $ x_2 $.\\\indent Simultaneously, $ x_1 $ and $ x_2 $ should satisfy the following constraints:
\begin{equation} {\omega _{x_1}}^T{R_{x_1x_1}}{\omega _{x_1}} = {\omega _{x_2}}^T{R_{x_2x_2}}{\omega _{x_2}} = 1, \end{equation}
where ${R_{x_1x_1}} = x_1{x_1^T}$ and ${R_{x_2x_2}} = x_2{x_2^T}$. \\\indent By solving equation (1) using the algorithm of Lagrange multipliers, we obtain the following relationship [17]:
\begin{equation} \left[ \begin{array}{l}
 0{\rm{        }} \\
 {R_{x_2x_1}}{\rm{     }} \\
 \end{array} \right.\left. \begin{array}{l}
 {{\rm{R}}_{x_1x_2}} \\
 {\rm{0}} \\
 \end{array} \right]\omega {\rm{ = }}\mu \left[ \begin{array}{l}
 {{\rm{R}}_{x_1x_1}}{\rm{        }} \\
 {\rm{   0  }} \\
 \end{array} \right.\left. \begin{array}{l}
 0 \\
 {R_{x_2x_2}} \\
 \end{array} \right]\omega,
 \end{equation} \\ where $ \mu $ is the canonical correlation value and $\omega  = {[{\omega _{x_1}^T},{\omega _{x_2}^T}]^T}$ is the projected vector.
\subsection{Fundamentals of GCCA}
Let $ x_1 \in {R^{\;m}},x_2 \in {R^{\;p}} $ be two sets of variables as the entries. The correlation between the two random variables $x_1$ and $x_2$ is
\begin{equation}
J({\omega _{{x_1}}},{\omega _{{x_2}}}) = \frac{{{\omega _{{x_1}}}^T{R_{{{{x_1}}}{{{x_2}}}}}{\omega _{{x_2}}}}}{{{{({\omega _{{x_1}}}^T{R_{{{{x_1}}}{{{x_1}}}}}{\omega _{{x_1}}}{\omega _{{x_2}}}^T{R_{{{{x_2}}}{{{x_2}}}}}{\omega _{{x_2}}})}^{\frac{1}{2}}}}}.
\end{equation}
 Then, GCCA is mathematically expressed as:
 \begin{equation} \mathop {\arg \max }\limits_{{\omega _{x_1}},{\omega _{x_2}} }  \sum\limits_{\scriptstyle k,l = 1 \hfill \atop
  \scriptstyle {\rm{ }}k \ne l \hfill}^2 c_{kl} g(J({\omega _{{x_k}}},{\omega _{{x_l}}})) \end{equation}\\
  subject to \begin{equation}
{\omega _{{x_1}}}^T{R_{{x_1}{x_1}}}{\omega _{{x_1}}} = {\omega _{{x_2}}}^T{R_{{x_2}{x_2}}}{\omega _{{x_2}}} = 1,
 \end{equation}\\
 where $c_{kl}$ is the coefficient of a design matrix and \emph{g} stands for the identity, the absolute value, or the square function [15]. The two vectors $\omega _{{x_1}}$ and $\omega _{{x_2}}$ maximizing the criterion function in equation (5), are called GCCA. The solution to equation (5) is obtained by solving the Lagrange multipliers as follows [14]:
 \begin{equation}
 \left[ {\begin{array}{*{20}{c}}
   0  \\
   {{R_{{x_2}{x_1}}}}  \\
\end{array}} \right.\left. {\begin{array}{*{20}{c}}
   {{\rm{    }}{R_{{x_1}{x_2}}}}  \\
   {\rm{0}}  \\
\end{array}} \right]\omega {\rm{ = }}\eta \left[ {\begin{array}{*{20}{c}}
   {{S_{{x_1}{x_1}}}}  \\
   {\rm{0}}  \\
\end{array}} \right.\left. {\begin{array}{*{20}{c}}
   0  \\
   {{\rm{    }}{S_{{x_2}{x_2}}}}  \\
\end{array}} \right]\omega
\end{equation}
where ${S_{{x_1}{x_1}}}$ and ${S_{{x_2}{x_2}}}$ denote the within-class scatter matrix of training samples in feature space ${x_1}$ and ${x_2}$; $\eta$ is the canonical correlation value and $\omega  = {[{\omega _{x_1}^T},{\omega _{x_2}^T}]^T}$ is the projected vector.
\subsection{Fundamentals of MCCA}
 Given $ P $ sets of real random variables ${x_1},{\rm{ }}{x_2}, \cdots {x_P}$ with the dimensions of ${m_1},{\rm{ }}{m_2}, \cdots {m_P}$. The objective of MCCA is to find $\omega  = {[{\omega _{x_1}}^T,{\omega _{x_2}}^T \cdots {\omega _{x_P}}^T]^T}$ which satisfies a similar requirement of CCA which is mathematically expressed as: \\
 \begin{equation} \mathop {\arg \max }\limits_{{\omega _1},{\omega _2} \cdots {\omega _P}}  \frac{1}{{P(P - 1)}}\sum\limits_{\scriptstyle k,l = 1 \hfill \atop
  \scriptstyle {\rm{ }}k \ne l \hfill}^P {{\omega _{x_k}}^T{R_{{x_k}{x_l}}}{\omega _{x_l}}} {\rm{  (}}k \ne l{\rm{)}} \end{equation}\\
  subject to \begin{equation} \sum\limits_{k = 1}^P {{\omega _{x_k}}^T{R_{{x_k}{x_k}}}{\omega _{x_k}}}  = P, \end{equation}\\
  where ${R_{{x_k}{x_l}}} = {x_k}{x_l}^T $. Equation (8) can be transformed into equation (10) by introducing Lagrange multipliers\\
  \begin{equation} \frac{1}{{P - 1}}(C - D)\omega  = \beta D\omega, \end{equation} \\ where \\
\begin{equation}\ C = \left[ {\begin{array}{*{20}{c}}
   {{x_1}{x_1}^T} &  \ldots  & {{x_1}{x_P}^T}  \\
    \vdots  &  \ddots  &  \vdots   \\
   {{x_P}{x_1}^T} &  \cdots  & {{x_P}{x_P}^T}  \\
\end{array}} \right]
 \end{equation} \\
 \begin{equation}\ D = \left[ {\begin{array}{*{20}{c}}
   {{x_1}{x_1}^T} &  \ldots  & 0  \\
    \vdots  &  \ddots  &  \vdots   \\
   0 &  \cdots  & {{x_P}{x_P}^T}  \\
\end{array}} \right]
\end{equation} \\with $ \beta $ being the multiple canonical correlation value. The value of $ \beta $ can be calculated by the generalized eigenvalue (GEV) method.
\subsection{The Formulation of LMCCA}
In this subsection, we propose the method of LMCCA to improve the performance of information fusion. The advantages of LMCCA for information fusion rest on the following facts: 1) the correlation among the variables in multiple channels is taken as the metric of their similarity; 2) the class information is considered by LMCCA. \\\indent It is known that Linear Discriminant Analysis (LDA) [21-22] explores the projected vectors in the underlying space that best discriminate among classes (rather than those that best describe the data) and creates a linear combination of those which yields the largest mean differences between the desired classes. Mathematically, the purpose is to maximize the between-class scatter matrix and minimize the within-class scatter matrix simultaneously. In light of this purpose, LMCCA is proposed to improve the multiple canonical correlation criterion function by introducing the within-class measure information of the training samples to the multiple feature space. \\\indent

Given \emph{$ P $} sets of zero-mean random variables ${x_1} \in {R^{{m_1}}},{x_2} \in {R^{{m_2}}}, \cdots {x_P} \in {R^{{m_P}}}$ for \emph{c} classes and $Q = {m_1} + {m_2} +  \cdots {m_P} $. Let $ {S{\omega_{{x_t}}}} (t=1,2,...P) $ denotes the within-class scatter matrix of training samples in different features space, i.e. \\
\begin{equation}\ S{\omega_{{x_t}}} = \sum\limits_{i = 1}^c {p({\omega _i})\left[ {\sum\limits_{j = 1}^{{l_i}} {\frac{1}{{{l_i}}}({x_t}_{ij} - {m^{{x_t}}}_i){{({x_t}_{ij} - {m^{{x_t}}}_i)}^T}} } \right]} \end{equation}
where $ \ {x_t}_{ij} \in {x_t} \ $ $(t=1,2,...P)$ 
denotes the \textit{j}th training sample in class \textit{i} of the feature set $t$; $ p({\omega _i})$ is the prior probability of class \textit{i}; $ \textit{l}_\textit{i}$ is the number of training samples in class \textit{i}; and $ {m^{x_t}}_i $, 
 are the mean vectors of training samples in class \textit{i} with the feature space $t$, respectively. \\\indent
In order to reach maximum correlation when the projected vectors minimize the within-class scatter matrix, the criterion function of LMCCA is given as: \\
\begin{equation}
\mathop {\arg \max }\limits_{{\omega _{{x_1}}},{\omega _{{x_2}}} \cdots {\omega _{{x_P}}}} \frac{1}{{P(P - 1)}}\sum\limits_{\begin{array}{*{20}{c}}
   {k,l}  \\
   {k \ne l{\rm{ }}}  \\
\end{array}}^{{P}} {{\omega _{x_k}}^T{R_{x_kx_l}}{\omega _{x_l}}(k \ne l)}
\end{equation}\\
 subject to \begin{equation} \sum\limits_{{k=1}}^{P} {{\omega_{{x_k}}}^TS{\omega_{{x_k}}}{\omega_{{x_k}}}}  = P \end{equation}\\ where
${R_{x_kx_l}} = x_k{x_l^T}$ denotes the cross-correlation matrix of $x_k$ and $x_l$.\\\indent
Equation (14) is further written as:
\begin {equation}\mathop {\arg \max}\limits_{{\omega _{x_1}},{\omega _{x_2}} \cdots {\omega _{x_P}}}  \frac{1}{{(P - 1)}}\frac{{\sum\limits_{\scriptstyle k,l\hfill \atop
  \scriptstyle {\rm{ }}k \ne l \hfill}^{P} {{\omega _{x_k}}^T{R_{x_kx_l}}{\omega _{x_l}}} {\rm{  (}}k \ne l{\rm{)}}}}{{\sum\limits_{k = {1}}^{P} {{\omega_{{x_k}}}^TS{\omega_{x_k}}{\omega_{{x_k}}}} }} \end{equation} \\\indent As ${\omega _{x_t}}$ satisfies constraint (15), we apply the Lagrange multiplier to transform equation (16) into:
\begin{equation} \begin{array}{l}
 J({\omega _{x_1}},{\omega _{x_2}},...{\omega _{x_P}}) = \frac{1}{{P - 1}}\sum\limits_{\scriptstyle k,l \hfill \atop
  \scriptstyle {\rm{ }}k \ne l \hfill}^{P} {{\omega _{x_k}}^T{R_{x_kx_l}}{\omega _{x_l}}} {\rm{  (}}k \ne l{\rm{)}} \\
 {\rm{~~~~~~~~~~~~~~~~~~~}} - \frac{\lambda}{2} {\rm{(}}\sum\limits_{k = 1}^{P} {{\omega _{x_k}}^TS{\omega_{x_k}}{\omega _{x_k}}}  - P{\rm{)}} \\
 \end{array} \end{equation}
Let
\begin{small}
\begin{equation}{\frac{{\partial J({\omega _{{x_1}}},{\omega _{{x_2}}},...{\omega _{{x_p}}})}}{{\partial {\omega _{{x_t}}}}} =\frac{1}{{P - 1}}\sum\limits_{l = 1,\\\ l \ne t}^P {{R_{{x_t}{x_l}}}{\omega _{{x_l}}}}  - \lambda S{\omega _{{x_t}}}{\omega _{{x_t}}}}=0
\end{equation} \end{small}
According to (18), equation (16) is further written as: \\
 \begin{equation} \frac{1}{{P - 1}}E\omega  = \lambda F\omega \end{equation} \\ where
 \begin{equation} E = \left[ {\begin{array}{*{20}{c}}
0&{{R_{{x_1}{x_2}}}}&{{R_{{x_1}{x_3}}} \ldots }&{{R_{{x_1}{x_P}}}}\\
{{R_{{x_2}{x_1}}}}&0&{{R_{{x_2}{x_3}}} \ldots }&{{R_{{x_2}{x_P}}}}\\
{}&{}& \vdots &{}\\
{{R_{{x_P}{x_1}}}}&{{R_{{x_P}{x_2}}}}&{{R_{{x_P}{x_3}}} \ldots }&0
\end{array}} \right] \end{equation}
\begin{equation}\ F = \left[ {\begin{array}{*{20}{c}}
{S{\omega_{{x_1}}}}&0&{0 \ldots }&0\\
0&{S{\omega_{{x_2}}}}&{0 \ldots }&0\\
{}&{}& \vdots &{}\\
0&0&{0 \ldots }&{S{\omega_{{x_P}}}}
\end{array}} \right]        \end{equation}
\\\indent It is further written as:
\begin{equation} \frac{1}{{P - 1}}(G - F)\omega  = \lambda F\omega \end{equation}
where
\begin{equation}
{G = \left[ {\begin{array}{*{20}{c}}
   {S{\omega _{{x_1}}}} & {{R_{{x_1}{x_2}}}} & {{R_{{x_1}{x_3}}} \ldots } & {{R_{{x_1}{x_P}}}}  \\
   {{R_{{x_2}{x_1}}}} & {S{\omega _{{x_2}}}} & {{R_{{x_2}{x_3}}} \ldots } & {{R_{{x_2}{x_P}}}}  \\
   {} & {{\rm{             }} \vdots } & {} & {}  \\
   {{R_{{x_P}{x_1}}}} & {{R_{{x_P}{x_2}}}} & {{R_{{x_P}{x_3}}} \ldots } & {S{\omega _{{x_P}}}}  \\
\end{array}} \right]}
\end{equation}
\begin{equation} \omega  = {[{\omega _1}^T,{\omega _2}^T, \cdots {\omega _P}^T]^T}\end{equation}
\\\indent Based on equations (18)--(24), the value of ${\lambda}$ plays a decisive role in evaluating the relationship between cross-correlation and within-class matrixes. When ${\lambda}>0$, the corresponding projected vector contributes positively to the discriminative power in classification. On the other hand, the projected vector leads to reducing the discriminative power in classification with the non-positive value of ${\lambda}$. Therefore, the solution to equation (22) is to find the eigenvectors associated to the positive eigenvalues.\\\indent From studying the property of within-class scatter matrix $ {S{\omega_{{x_1}}}}, {S{\omega_{{x_2}}}}...{S{\omega_{{x_P}}}} $, an important characteristic of the proposed method is discovered: although the within-class scatter information of training samples is introduced to LMCCA, the number of projected dimension \emph{d} corresponding to the optimal recognition accuracy satisfies the following relation
\begin{equation} \ d \le Q\ \end{equation}\\
where $Q$ is the number of total feature dimensions.\\\indent
The derivation of equation (25) is given in Appendix \emph{A}. Based on (25), the optimally projected dimension in LMCCA has no relation to the number of training samples, a property especially significant when dealing with big data problems. This immediately leads to an algorithm to solve LMCCA's problem with \textit{d} serial eigenvectors expressed as follows:
\begin{equation}
X = {\left[ {\begin{array}{*{20}{c}}
   {{\omega _{{x_1},d}}} & 0 & {0 \ldots } & 0  \\
   0 & {{\omega _{{x_2},d}}} & {0 \ldots } & 0  \\
   {} & {} &  \vdots  & {}  \\
   0 & 0 & {0 \ldots } & {{\omega _{{x_P},d}}}  \\
\end{array}} \right]^T}\left( {\begin{array}{*{20}{c}}
   {{x_1}}  \\
   {{x_2}}  \\
    \vdots   \\
   {{x_P}}  \\
\end{array}} \right){\rm{ }},
\end{equation}
where $\omega _{{x_t},d} (t=1,2,...P)$ denotes \textit{d} serial eigenvectors and $P$ is the number of features.\\\indent From the above mathematical analysis, the discrimination power of the LMCCA effectively models the relationship among multiple information sources. We then study the complexity of LMCCA and related algorithms. Without loss of generality, we select a total of $N$ training samples with \emph{$ P $} sets of variables ${x_1} \in {R^{{m_1}}},{x_2} \in {R^{{m_2}}}, \cdots {x_P} \in {R^{{m_P}}}$ for \emph{c} classes and $Q = {m_1} + {m_2} +  \cdots {m_P} $. Then, the complexity of different algorithms is given in Table I, where
$ e = \max ({m_i},{m_j})(i,j \in (1,2,...P))$ and $k = \max ({m_1},{m_2},...{m_P})$.\\\indent
\vspace*{-10pt}
\begin{table}[h]
\small
\renewcommand{\arraystretch}{1.3}
\caption{\normalsize{The complexity of different algorithms}}
\setlength{\abovecaptionskip}{0pt}
\setlength{\belowcaptionskip}{10pt}
\centering
\tabcolsep 0.073in
\begin{tabular}{cc}
\hline
Algorithm & Complexity\\
\hline
CCA & $O(N \times e^2) +O(e^3)$\\
GCCA &$O(N \times e^2) +O(e^3)+ O(N \times (m_1+m_2)) $\\
MCCA &$O(N \times k^2) +O(k^3)$\\
LMCCA &$O(N \times k^2) +O(k^3)+ O(N \times Q) $\\
\hline
\end{tabular}
\end{table}\\

From Table I, compared with MCCA, LMCCA introduced an extra computational complexity of $O(N \times Q)$. However, the gain in performance over MCCA in the experiments (4.5\% for handwritten digits to 12\% for emotion recognition) as shown in Section IV outweighs the extra cost in computation.
\subsection{The Relationship between CCA, MCCA, GCCA, LMCCA}
The derivation of LMCCA shows that CCA, MCCA and GCCA are special cases of LMCCA as demonstrated below:\\

\textbf{(1)} When \emph{P}=2, (20) and (21) turn into the following form
\begin{equation}
E = \left[ {\begin{array}{*{20}{c}}
   0  \\
   {{R_{{x_2}{x_1}}}}  \\
\end{array}} \right.\left. {\begin{array}{*{20}{c}}
   {{R_{{x_1}{x_2}}}}  \\
   {\rm{0}}  \\
\end{array}} \right],F = \left[ {\begin{array}{*{20}{c}}
   {S{\omega _{{x_1}}}}  \\
   0  \\
\end{array}} \right.\left. {\begin{array}{*{20}{c}}
   0  \\
   {S{\omega _{{x_2}}}}  \\
\end{array}} \right]
 \end{equation}\\\indent
Thus, LMCCA is reduced to the method of GCCA [10].\\

\textbf{(2)} When $ S{\omega _{x_1}}$ ... $ S{\omega _{x_P}}$ are autocorrelation matrixes, i.e.

\begin{equation} \ S{\omega_{x_1}} \to {R_{{x_1}{x_1}}},\;...,S{\omega_{{x_P}}} \to {R_{{x_P}{x_P}}}, \end{equation}
(20) and (21) turn into the following form
\begin{equation} \ E = \left[ {\left( {\begin{array}{*{20}{c}}
   0 &  \ldots  & {{R_{{x_1}{x_P}}}}  \\
    \vdots  &  \ddots  &  \vdots   \\
   {{R_{{x_P}{x_1}}}} &  \cdots  & 0  \\
\end{array}} \right)} \right] \end{equation}

\begin{equation} \ F = \left[ {\left( {\begin{array}{*{20}{c}}
{{R_{{x_1}{x_1}}}}& \ldots &0\\
 \vdots & \ddots & \vdots \\
0& \cdots &{{R_{{x_P}{x_P}}}}
\end{array}} \right)} \right]
 \end{equation}\\\indent
LMCCA is reduced to the method of MCCA.\\

\textbf{(3)} When \emph{P}=2 and $ S{\omega _{x_1}}, S{\omega _{x_2}}$ are autocorrelation matrixes,
\begin{equation} \ S{\omega_{x_1}} \to {R_{{x_1}{x_1}}}, S{\omega_{{x_2}}} \to {R_{{x_2}{x_2}}}, \end{equation}
(20) and (21) turn into the following form
\begin{equation}
E = \left[ {\begin{array}{*{20}{c}}
   0  \\
   {{R_{{x_2}{x_1}}}}  \\
\end{array}} \right.\left. {\begin{array}{*{20}{c}}
   {{R_{{x_1}{x_2}}}}  \\
   {\rm{0}}  \\
\end{array}} \right],F = \left[ {\begin{array}{*{20}{c}}
   {{R_{{x_1}{x_1}}}}  \\
   0  \\
\end{array}} \right.\left. {\begin{array}{*{20}{c}}
   0  \\
   {{R_{{x_2}{x_2}}}}  \\
\end{array}} \right]
\end{equation}\\\indent
 LMCCA is reduced to the method of CCA.
\section{Feature Extraction and Classification}
In this section, we present the classification method used with LMCCA and features used in the four application scenarios: handwritten digit recognition, face recognition, object recognition and human emotion recognition. Handwritten digit recognition is a typical field of application for automatic classification methods. Its main application areas fall in postal mail sorting, bank check processing and form data entry. Face recognition is a vitally important research area spanning multiple fields and disciplines. It is essential for effective communications and interactions among people with applications to bankcard identification, mugshot searching, surveillance systems, etc. Object recognition is a process for identifying a specific object in a digital image or video. It is popularly used in numerous applications such as video stabilization, automated vehicle parking systems, and cell counting in bioimaging. Human emotion has been playing an important role in our daily social interactions and activities. In this paper, the study of audio and visual information for emotion recognition serves as an example in the performance evaluation of the proposed method.\\
\subsection{Feature Extraction}
To demonstrate the effectiveness of LMCCA, different features are extracted as follow:
\subsubsection{Handwritten Digit Features}
There are many energy based descriptor available to handwritten digit features extraction such as Harris, Contrast etc [69]. In this paper, we extracted three sets of features for handwritten digit recognition [23-24]:\\
a) 24-dimensional: the mean of the digit images transformed by the Gabor filter.\\
b) 24-dimensional: the standard deviation of the digit images transformed by the Gabor filter.\\
c) 36-dimensional: Zernike moment feature [23].
\subsubsection{Face Features}
In the evaluation of the proposed method for face recognition, we extracted the following three kinds of features [25-27]:\\
d) 36-dimensional: the histogram of oriented gradient (HOG) of the face images.\\
e) 33-dimensional: the local binary Patterns (LBP) of the face images.\\
f) 48-dimensional: Gabor transformation features with the face images.
\subsubsection{Object Features}
In terms of object recognition, GIST [44], Pyramid of Histogram of Oriented Gradients (PHOG) [45] and Local Binary Patterns (LBP) [46] are extracted as features:\\
g) 20-dimensional: GIST of the object images.\\
h) 59-dimensional: PHOG of the object images.\\
i) 40-dimensional: LBP of the object images.
\subsubsection{Audio Features for Emotion Recognition}
In this work three sets of audio features, Prosodic, MFCC and Formant Frequency (FF), are utilized to represent audio characteristics in emotion recognition [28-32]:\\
j) 25-dimensional: Prosodic features used in [33].\\
k) 65-dimensional: MFCC features (the mean, median, standard deviation, max, and range of the first 13 MFCC coefficients).\\
l) 15-dimensional: Formant Frequency features (the mean, median, standard deviation, max and min of the first three formant frequencies).
\subsubsection{Visual Features for Emotion Recognition}
For facial features, we calculated the mean, standard deviation and median of the magnitude of the facial images transformed by the Gabor filter. The designed Gabor filter bank consists of filters in 4 scales and 6 orientations [34-37].\\
m) 24-dimensional: the mean of the facial images transformed by the Gabor filter.\\
n) 24-dimensional: the standard deviation of the facial images transformed by the Gabor filter.\\
o) 24-dimensional: the median of the facial images transformed by the Gabor filter.\\\indent

After that, a generic block diagram of the proposed fusion-recognition system is depicted as Fig. 1.\\
\centerline {\includegraphics[width=1.8in]{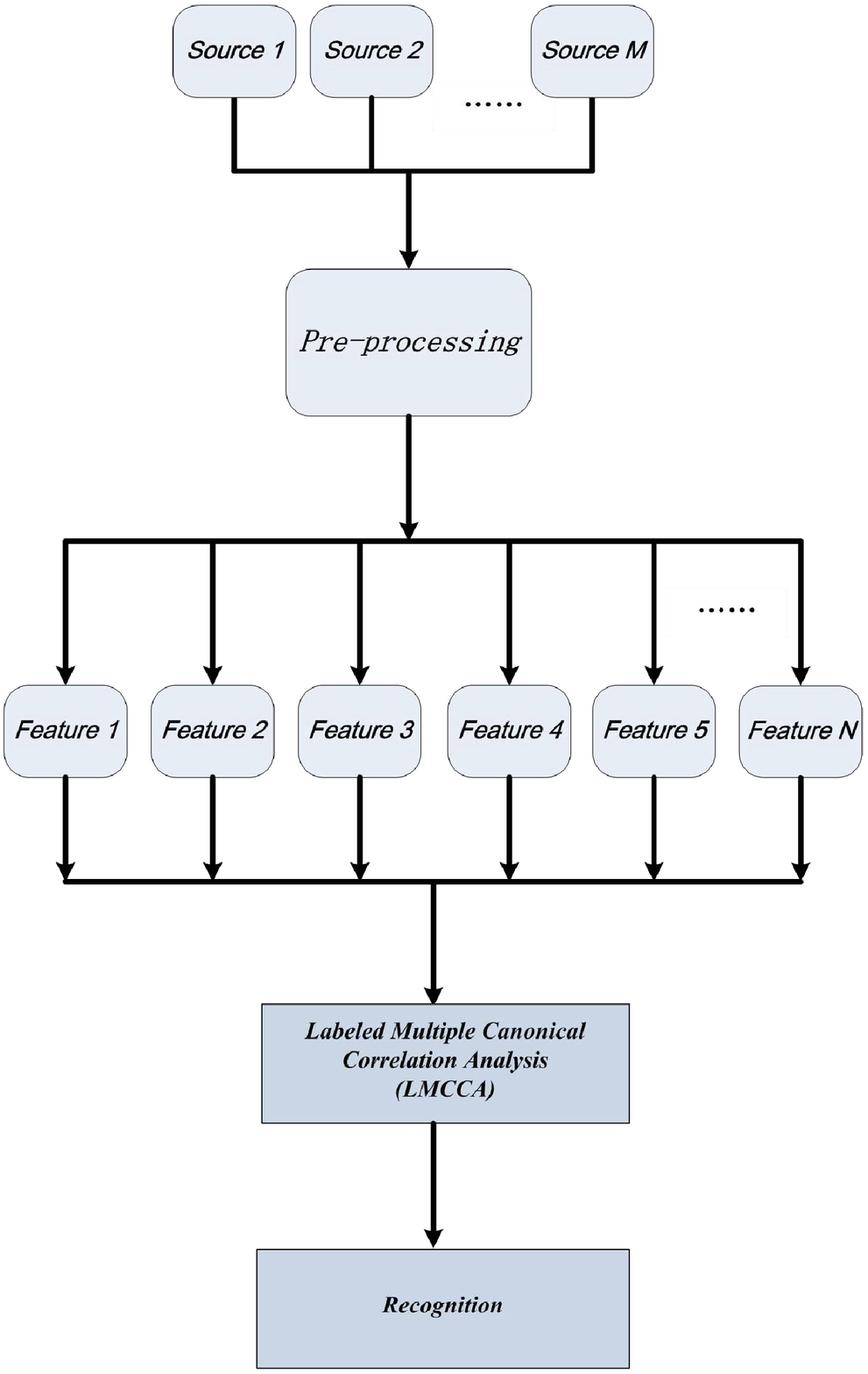}}\\  \centerline {Fig. 1 The block diagram of LMCCA fusion-recognition system}\\\indent
\subsection{Classification Method}
For the classification method, we use the algorithm proposed in [38]. The classification algorithm is summarized below:\\
\textbf{1.} Given two sets of features, represented by feature matrices
\begin{equation} \ {X^1} = [{x^1}_1,{x^1}_2,{x^1}_3,...{x^1}_d], \end{equation}
and
\begin{equation} \ {X^2} = [{x^2}_1,{x^2}_2,{x^2}_3,...{x^2}_d]. \end{equation}.\\
\textbf{2.} $dist[{X^1}, {X^2}]$ is defined as
\begin{equation} \ dist[{X^1}, {X^2}] = \sum\limits_{j = 1}^d {{{\left\| {{x^1}_j - {x^2}_j} \right\|}_2}}, \end{equation}
where ${\left\| {a - b} \right\|_2}$ denotes the Euclidean distance between the two vectors \emph{a} and \emph{b}. \\
\textbf{3.} Let the feature matrices of the \emph{N} training samples be ${F_1},{F_2},...{F_N}$ and each sample belongs to some class
${C_i}$ $ (i = 1,2...c )$.\\
\textbf{4.} For a given test sample \emph{I}, if
\begin{equation} \;dist[I,{F_l}] = \mathop {\min}\limits_j dist[I,{F_j}] (j = 1,2...N),\end{equation}
and
\begin{equation}
class{\rm{ }}of{\rm{ \{ }}{F_l}\}  \in {C_i}
,\end{equation}
the resulting decision is $I \in {C_i}$. \\\indent

Finally, in summary, the information fusion algorithm based on LMCCA is given below:\\
\textbf{Step 1.} Extract information from multimodal sources to form the training sample spaces.\\
\textbf{Step 2.} Convert the extracted features into the normalized form and compute the matrixes $F$ and $G$.\\
\textbf{Step 3.} Compute the eigenvalues in the matrix $\lambda$ and eigenvectors in the matrix $\omega$ of equation (22).\\
\textbf{Step 4.} Obtain the fused information expression from equation (26), which is used for classification.
\section{PERFORMANCE EVALUATION AND ANALYSIS}
To evaluate the effectiveness of the proposed solution, we conduct experiments on Mixed National Institute of Standards and Technology (MNIST) handwritten digit database, ORL face database, Caltech 101 object database, Ryerson Multimedia Lab (RML) [33] and eNTERFACE (eNT) [39] emotion database.
\subsection{Performance of Handwritten Digit Recognition}
The MNIST database, or modified NIST database, is constructed out of the original NIST database. All the digits are size normalized, and centered in a fixed size image where the center of gravity of the intensity lies at the center of the image with 28 $\times$ 28 pixels, where the pixels take on binary values. Example images from MNIST database are shown in Fig. 2.\\
\centerline {\includegraphics[width=1.8in]{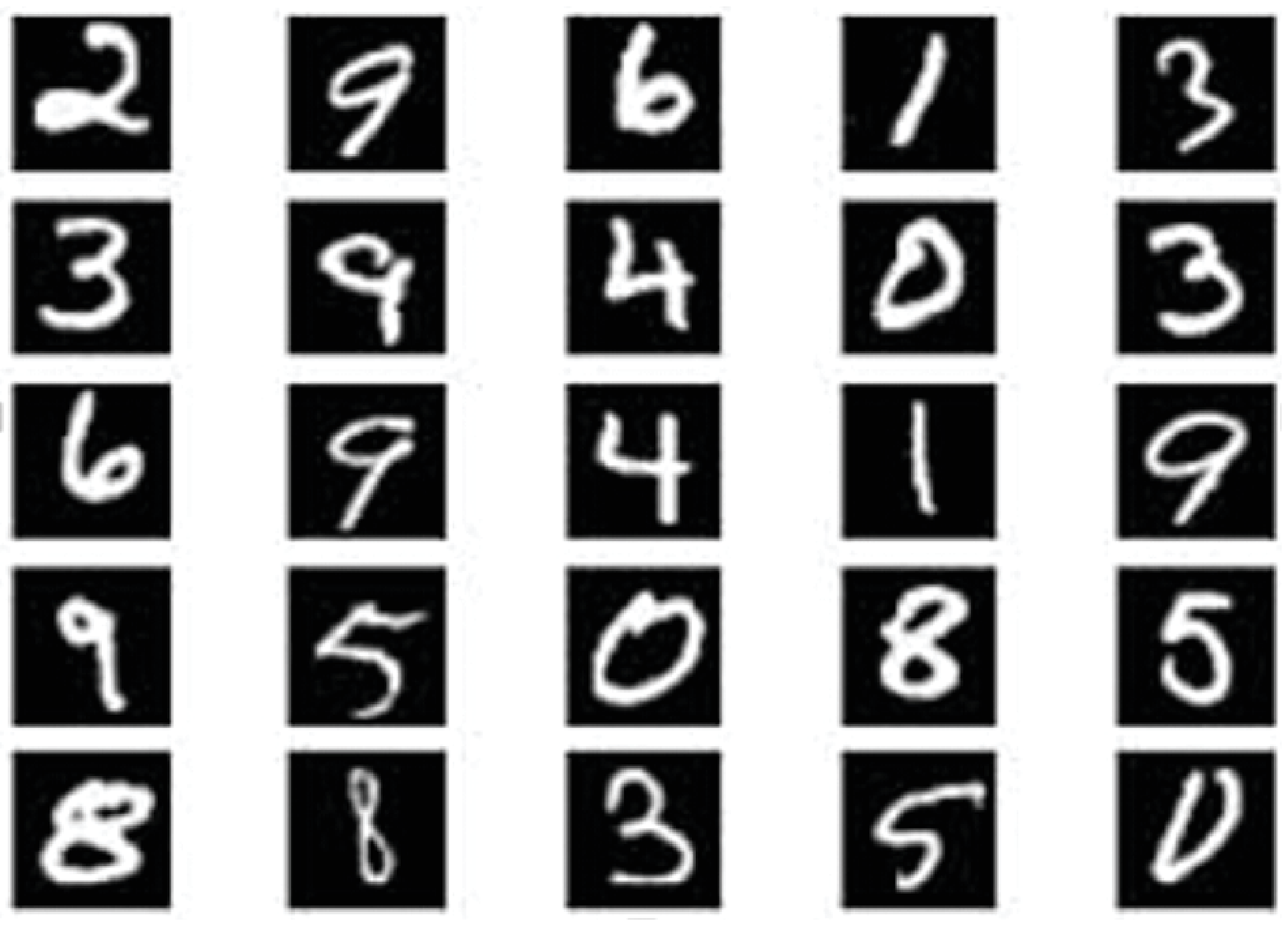}}\\  \centerline {Fig. 2  Example images from the MNIST database}\\\indent

In the experiments, 3000 handwritten digital samples from 10 classes, or digits (from 0 to 9), are selected. We divide the samples into training and testing subsets, each containing 1500 samples. The performance of using mean, standard deviation and Zernike moment features is first evaluated shown in TABLE II. The recognition accuracy is calculated as the ratio of the number of correctly classified samples over the total number of testing samples.
\vspace*{-10pt}
\begin{table}[h]
\small
\renewcommand{\arraystretch}{1.3}
\caption{\normalsize{Results of handwritten digit recognition with a single feature}}
\setlength{\abovecaptionskip}{0pt}
\setlength{\belowcaptionskip}{10pt}
\centering
\tabcolsep 0.073in
\begin{tabular}{cc}
\hline
Single Feature & Recognition Accuracy\\
\hline
Mean &49.13\%\\
Standard Deviation &52.60\%\\
Zernike &70.20\%\\
\hline
\end{tabular}
\end{table}\\

TABLE II suggests we use the standard deviation (\textbf{52.60\%}) and Zernike moment (\textbf{70.20\%}) features which perform the best individually, in CCA and GCCA which only take two sets of features.
Then the performance of LMCCA with those of serial fusion, CCA, GCCA, and MCCA is ready to be compared. The overall recognition rates are given in Fig. 3, showing that the proposed LMCCA outperforms the other methods. In addition, from this figure, it is clear that the application of LMCCA achieves the best performance when the projected dimension \emph{d} is equal to 23, which is less than the total number of features \emph{Q}=24+24+36=84, confirming nicely with the mathematical proof in Section II and Appendix A.\\\indent We further implemented the experiments using deep learning based methods [42-43] and the parameters are given as follows:\\
\textbf{CNN}: iterations=550 and learning rate is 0.03.\\
\textbf{PCANet}: the filter size $k_1$=$k_2$=7 and the number of filters in each stage $L_1$=$L_2$=4.\\\indent
Table III summarizes their best performance and comparison with the SML methods. LMCCA shows the best performance on this dataset. \\
\centerline {\includegraphics[width=4.0in]{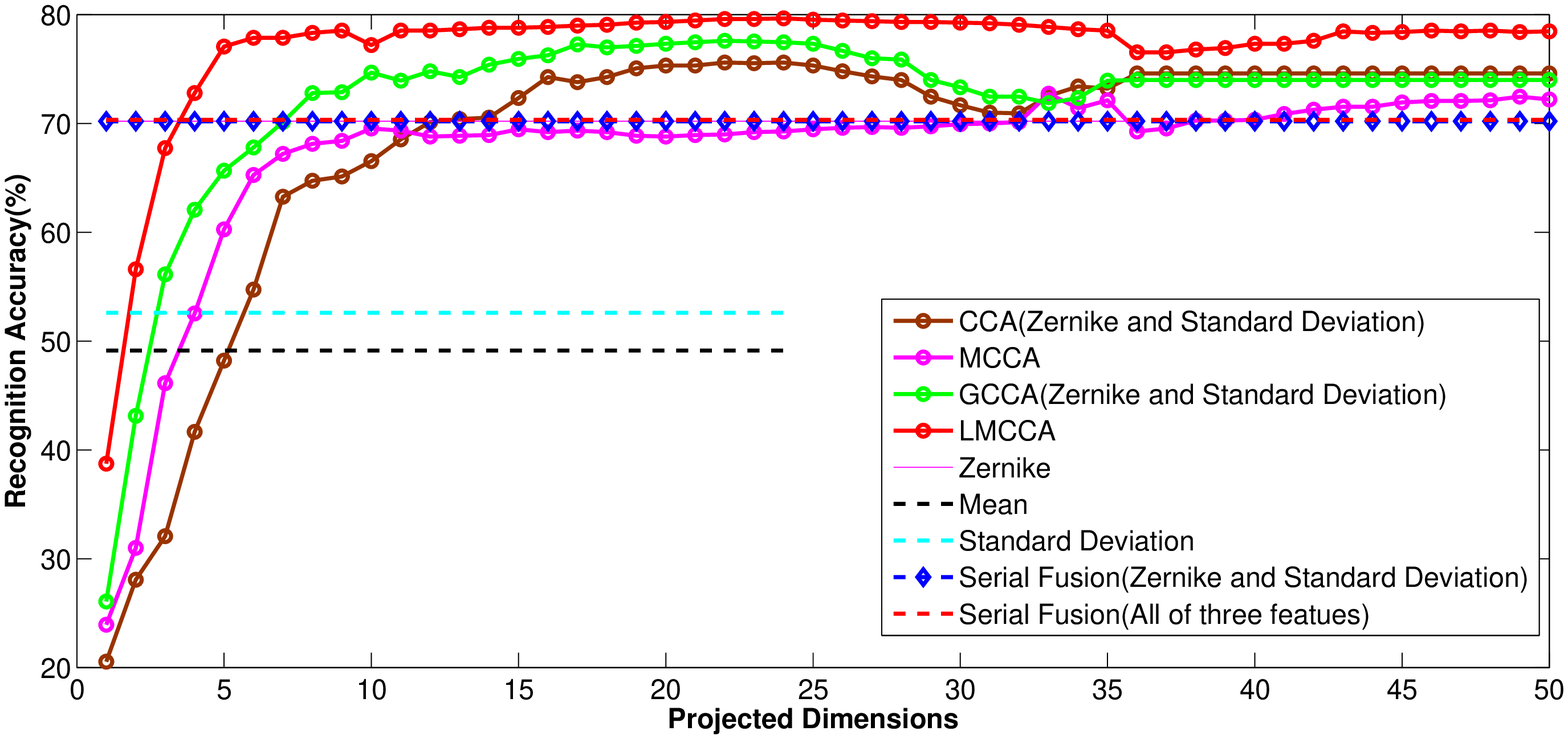}}\\  \centerline {Fig. 3 Experimental results of SML methods on MNIST database}
\vspace*{-10pt}
\begin{table}[h]
\small
\renewcommand{\arraystretch}{1.4}
\caption{\normalsize{Relation between optimal accuracy and dimension (Handwritten Digit)}}
\setlength{\abovecaptionskip}{0pt}
\setlength{\belowcaptionskip}{10pt}
\centering
\tabcolsep 0.07in
\begin{tabular}{ccc}
\hline
Method & Highest Accuracy & Dimension(Number)\\
\hline
Serial Fusion [9] &70.33\% & ---\\
CCA [11] &74.60\% & 36\\
GCCA [14] &75.53\% & 24\\
MCCA [19] &72.73\% & 33\\
CNN [42] &76.40\% & ---\\
PCANet [43] &79.20\% & ---\\
\textbf{LMCCA} &\textbf{79.96\%} & 23\\
\hline
\end{tabular}
\end{table}\\\indent
It is worth noting that both Zernike moment and standard deviation provide better individual performance, \textbf{70.20\%} and \textbf{56.20\%}, than that by mean alone, \textbf{49.13\%}, for the MNIST database. Since CCA and MCCA only analyze the correlation between different variables without considering discriminative representations, when it is used in multimodal information fusion and there is very different performance among features such as the MNIST database, there is no guarantee that using more features would achieve higher recognition accuracy.
\subsection{Performance of Face Recognition}
\fancyhead[RO]{http://www.cam-orl.co.uk}
The ORL database contains images from 40 individuals, each providing 10 different images. Each image is normalized and centered in a gray-level image with size 64 $\times$ 64, or 4096 pixels in total. Ten sample images of two subjects from the ORL database are shown in Fig. 4.\\
\centerline {\includegraphics[width=3.0in]{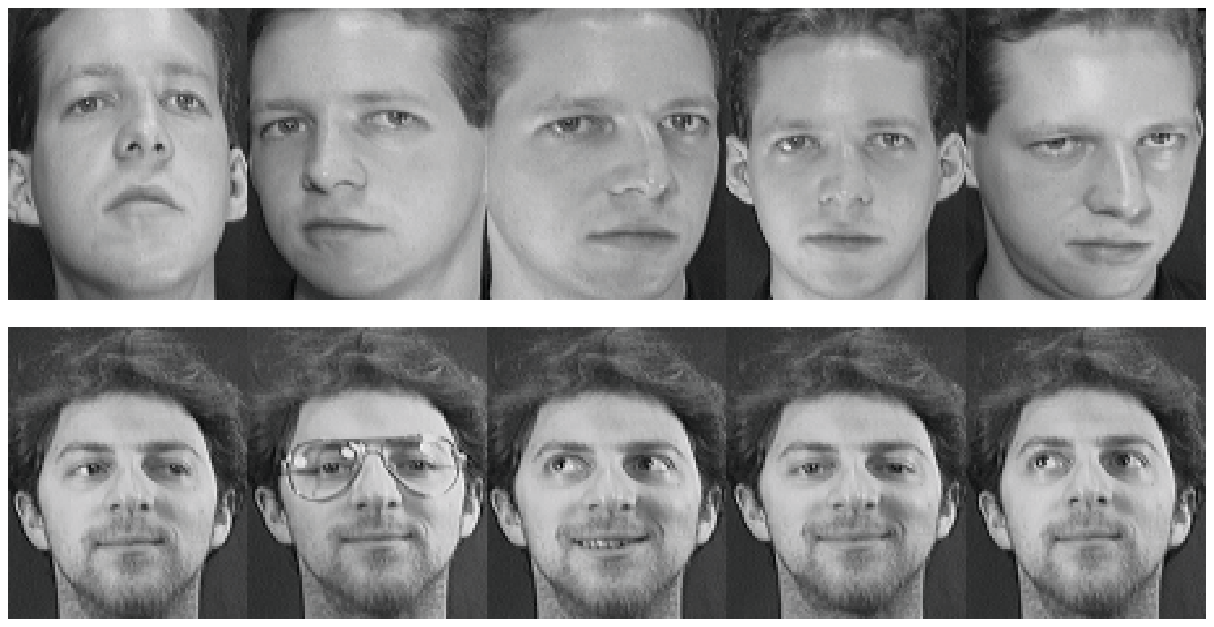}}\\  \centerline {Fig. 4 Images of two persons in the ORL database}\\\indent
\vspace*{-10pt}
\begin{table}[h]
\small
\renewcommand{\arraystretch}{1.3}
\caption{\normalsize{Results of face recognition with a single feature (ORL)}}
\setlength{\abovecaptionskip}{0pt}
\setlength{\belowcaptionskip}{10pt}
\centering
\tabcolsep 0.07in
\begin{tabular}{cc}
\hline
Single Feature & Recognition Accuracy\\
\hline
 HOG  &90.50\%\\
 LBP  &77.50\%\\
 Gabor &85.50\%\\
\hline
\end{tabular}
\end{table}\\\indent
The proposed algorithm is tested on the whole ORL database. The evaluation is based on cross-validation, where five images of each subject are chosen for training, while the remaining five images are used for testing. Thus, the training sample set size is 200 and the testing sample set size is 200. The performance of using HOG, LBP and Gabor features in a face image is shown in TABLE IV. From TABLE IV, it suggests we use the HOG (\textbf{90.50\%}) and Gabor (\textbf{85.50\%}) which provide the best individual performance as the input to CCA and GCCA.\\\indent 
The performance by the methods of CCA, GCCA, MCCA, and LMCCA is shown in Fig. 5. From the experimental results, clearly, LMCCA is more effective to handle the face recognition problem. Moreover, LMCCA achieves the optimal performance when the projected dimension \emph{d} is equal to 27, which is less than \emph{Q}=36+33+48=117. After that, comparison among serial fusion of the three features, CCA, GCCA, MCCA, discriminative sparse representation (DSR) [51], collaborative representation classification (CRC) [52], \emph{$l_1$}-regularized least squares (L1LS) [53], dual augmented lagrangian method (DALM) [54] and two DL methods [42-43] are presented. The parameters for the two DL methods are shown as follows:\\
\textbf{CNN}: iterations=800 and learning rate is 0.03.\\
\textbf{PCANet}: the filter size $k_1$=$k_2$=7 and the number of filters in each stage $L_1$=$L_2$=3.\\\indent
The optimal recognition accuracies by these methods are tabulated in TABLE V, showing that LMCCA outperforms the other methods.\\\indent
\centerline {\includegraphics[width=4.0in]{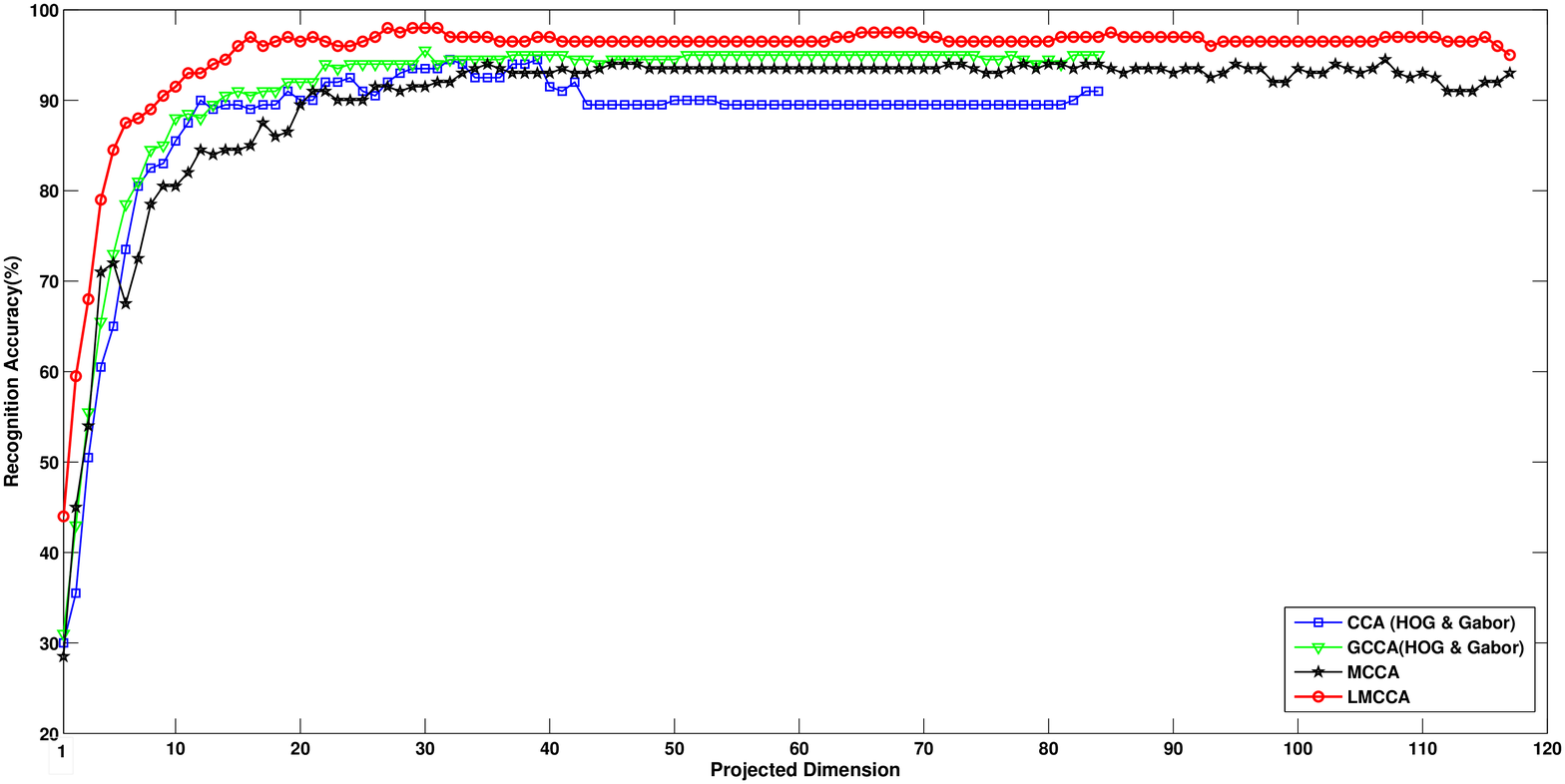}}\\  \centerline {Fig. 5 Face recognition experimental results on ORL database}\\\indent
\vspace*{-10pt}
\begin{table}[h]
\small
\renewcommand{\arraystretch}{1.4}
\caption{\normalsize{Relation between optimal accuracy and dimension (ORL)}}
\setlength{\abovecaptionskip}{0pt}
\setlength{\belowcaptionskip}{10pt}
\centering
\tabcolsep 0.07in
\begin{tabular}{ccc}
\hline
Method & Highest Accuracy & Dimension(Number)\\
\hline
Serial Fusion [9] &77.50\% & ---\\
CCA [11] &94.50\% & 32\\
GCCA [14] &95.50\% & 30\\
MCCA [19] &94.50\% & 72\\
DSR [51] &94.50\% & ---\\
CRC [52] &88.50\%& ---\\
L1LS [53] &92.50\%& ---\\
DALM [54] &90.00\%& ---\\
CNN [42] &76.00\% & ---\\
PCANet [43] &92.00\% & ---\\
\textbf{LMCCA} &\textbf{98.00\%} & 27\\
\hline
\end{tabular}
\end{table}
\subsection{Performance of Emotion Recognition}
The third and more detailed performance evaluation of the proposed method is conducted on human emotion recognition on RML and eNTERFACE (eNT) emotion database. The RML database consists of video samples of the six principal emotions  (angry, disgust, fear, surprise, sadness and happiness), performed by eight subjects speaking six different languages  (English, Mandarin, Urdu, Punjabi, Persian, and Italian). The frame rate for the videos is 30 fps with audio recorded at a sampling rate of 22050Hz. The image frame is 720 $\times$ 480 pixels, and the average size for face region is 112 $\times$ 96 pixels [41]. The eNT database contains video samples from 43 subjects, also expressing the six basic emotions, with a video frame rate of 25 fps and a sampling rate of 48000Hz for audio channel. The image frames have a size of 720 $\times$ 576 pixels, with the average size of the face region 260 $\times$ 300 pixels [41]. Example facial expression images from RML and eNT are shown in Fig. 6.\\\indent
\centerline {\includegraphics[width=3.0in]{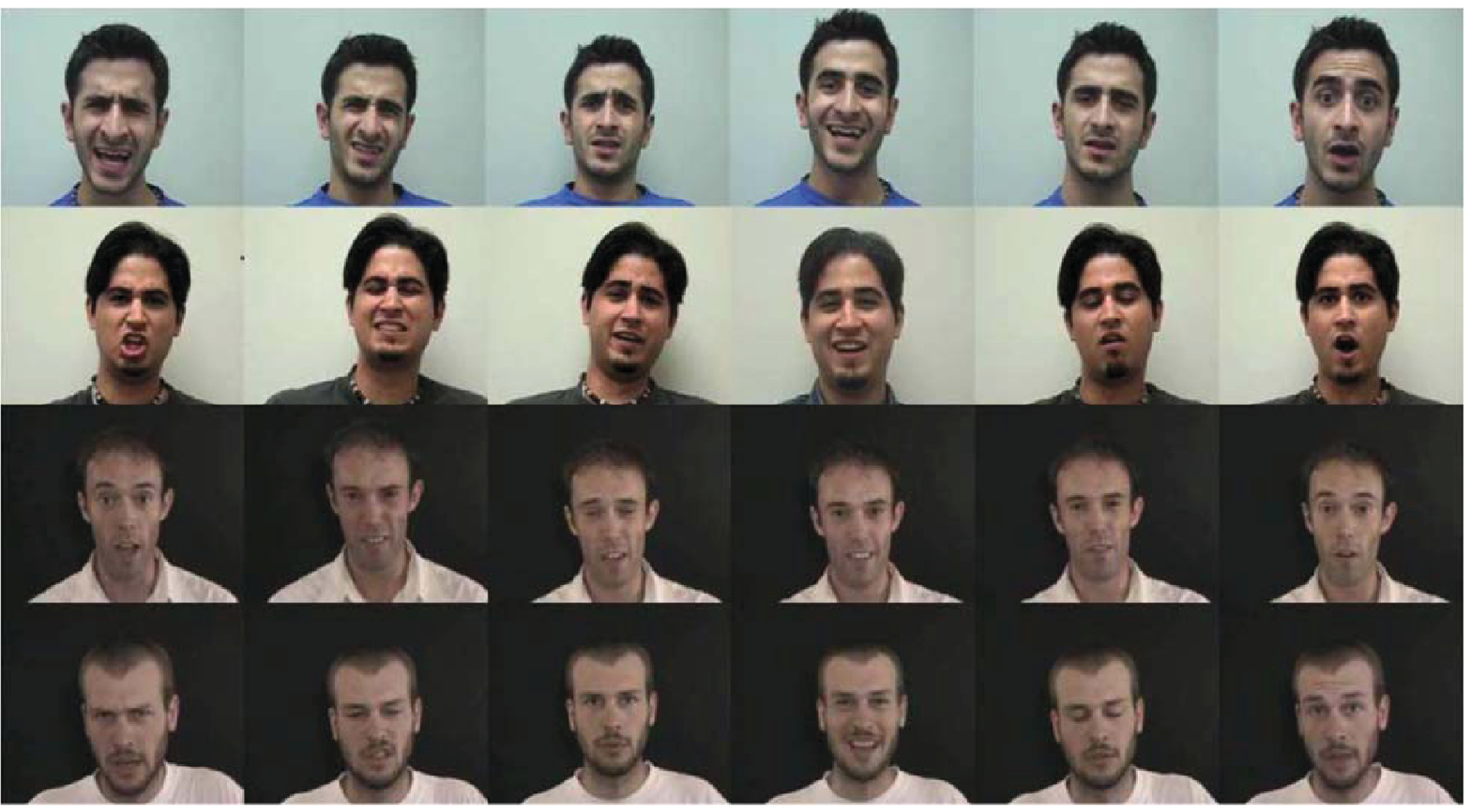}}\\  \leftline {Fig. 6 Example facial expression images
from the RML (the} {top two rows) and eNT (the bottom two rows) Databases}\\\indent

In the experiments, from each database, 456 video clips of the six basic emotions are selected for capturing the change of audio and visual information with respect to time simultaneously. The video clips from each database are randomly divided into 360 training and 96 testing samples, respectively. For benchmark purpose, the performance of using mean, standard deviation, median, Prosodic, MFCC and Formant Frequency features is first evaluated and tabulated in TABLE VI.\\\indent
\vspace*{-10pt}
\begin{table}[h]
\small
\renewcommand{\arraystretch}{1.4}
\caption{\normalsize{Results of Audiovisual emotion recognition on a single feature}}
\setlength{\abovecaptionskip}{0pt}
\setlength{\belowcaptionskip}{10pt}
\centering
\tabcolsep 0.073in
\begin{tabular}{cc}
\hline
Single Feature & Recognition Accuracy\\
\hline
Prosodic(RML) &45.83\%\\
MFCC(RML) &34.38\%\\
Formant Frequency(RML) &22.92\%\\
Mean(RML) &60.42\%\\
Standard Deviation(RML) &69.79\%\\
Median(RML) &57.29\%\\
Prosodic(eNT) &55.21\%\\
MFCC(eNT) &39.58\%\\
Formant Frequency(eNT) &31.25\%\\
Mean(eNT) &75.00\%\\
Standard Deviation(eNT) &80.21\%\\
Median(eNT) &72.92\%\\
\hline
\end{tabular}
\end{table}\\\indent
TABLE VI suggests we use the Prosodic (\textbf{45.83\%}, \textbf{55.21\%}) and standard deviation (\textbf{69.79\%}, \textbf{80.21\%}) features which perform the best individually, in CCA and GCCA which only take two sets of features. We also experiment on the method of serial fusion on RML and eNTERFACE database with all the six features, respectively. The experimental results are tabulated in TABLE VII. From TABLE VII, it demonstrates that there is no guarantee that serial fusion can achieve better recognition accuracy by using more features.\\

\vspace*{-10pt}
\begin{table}[h]
\small
\renewcommand{\arraystretch}{1.4}
\caption{\normalsize{Results of Audiovisual emotion recognition by serial fusion}}
\setlength{\abovecaptionskip}{0pt}
\setlength{\belowcaptionskip}{10pt}
\centering
\tabcolsep 0.07in
\begin{tabular}{cc}
\hline
Serial Fusion & Recognition Accuracy\\
\hline
 All of the six features(RML) &43.75\%\\
 All of the six features(eNT) &44.79\%\\
\hline
\end{tabular}
\end{table}
Then, the overall performance by LMCCA and that by CCA, GCCA and MCCA on RML and eNT datasets are plotted in Fig. 7 and Fig. 8, respectively. The results show that the discriminative power of the LMCCA provides more effective modeling for bimodal audiovisual information fusion. Again, LMCCA achieves the best performance when the projected dimension \emph{d} is equal to 9 (RML) and 28 (eNT) respectively, which is less than \emph{Q}=24+24+24+25+65+15=177, confirming nicely with the mathematical proof in Section II and Appendix A.\\\indent After that, we compare the performance of LMCCA with those of CCA, GCCA, MCCA and a hybrid DL model (HDLM) [55]. The relation between highest recognition accuracy and dimension using different methods is shown in TABLE VIII. It is observed from the table that LMCCA is competitive with HDLM on RML dataset, but performs remarkably better than the methods being compared on eNT.
\centerline {\includegraphics[width=3.8in]{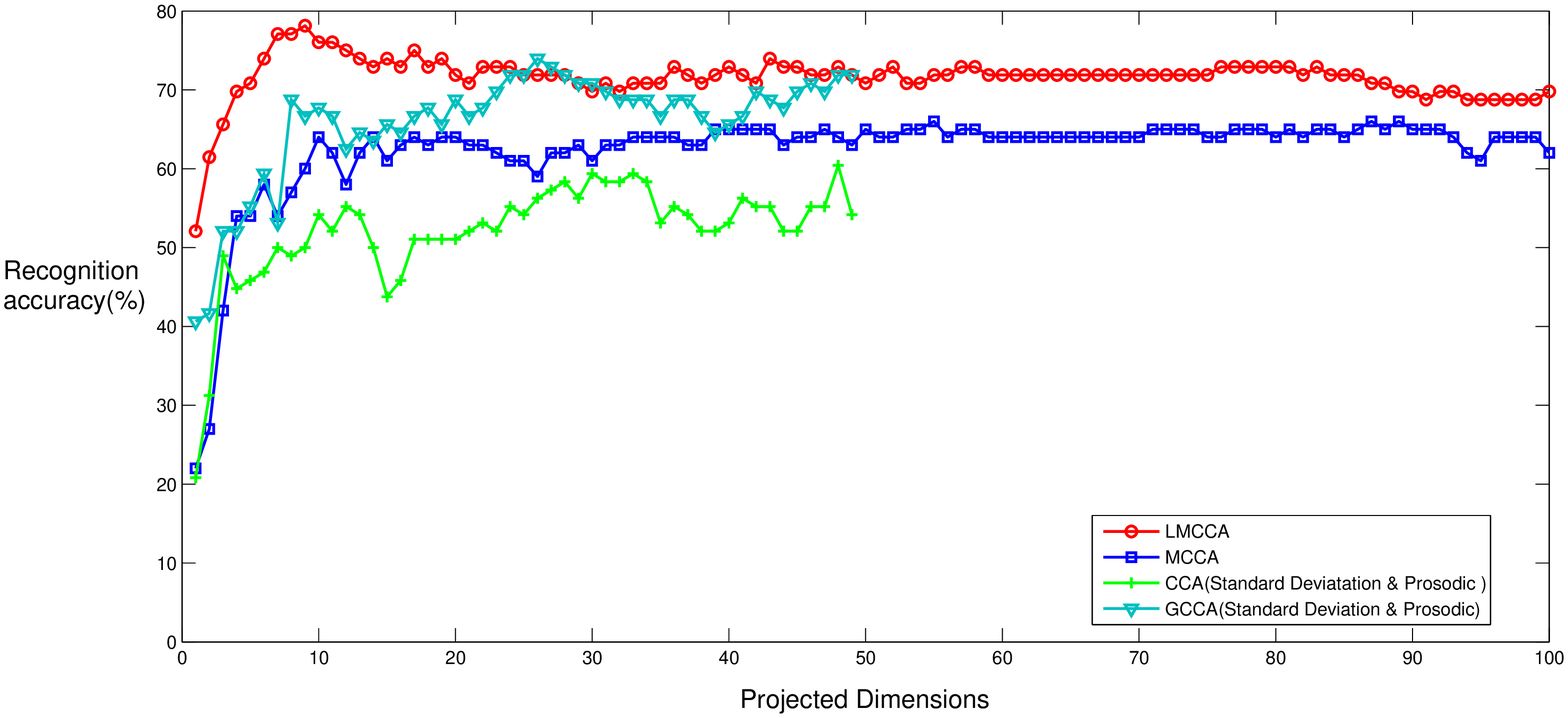}}\\  \centerline {Fig. 7 Experimental results on the RML database}
\centerline {\includegraphics[width=3.8in]{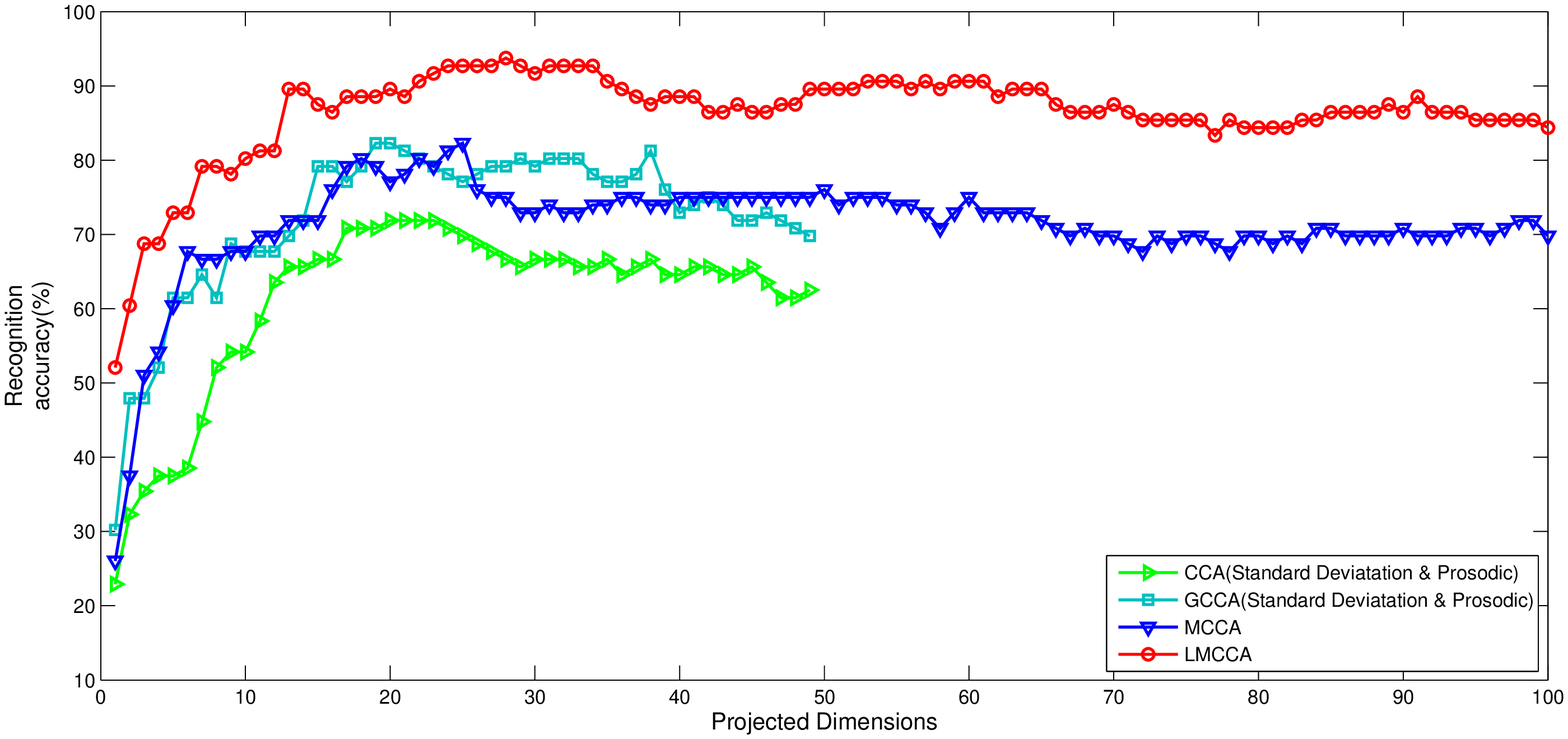}}\\  \centerline {Fig. 8 Experimental results on the eNT database}\\\indent
\vspace*{-10pt}
\begin{table}[h]
\small
\renewcommand{\arraystretch}{1.4}
\caption{\normalsize{Relation between optimal accuracy and dimension (Audiovisual)}}
\setlength{\abovecaptionskip}{0pt}
\setlength{\belowcaptionskip}{10pt}
\centering
\tabcolsep 0.07in
\begin{tabular}{ccc}
\hline
Method & Highest Accuracy & Dimension(Number)\\
\hline
\textbf{LMCCA(RML)} & \textbf{78.13\%} & 9\\
HDLM(RML) [55] &80.36\% &---\\
MCCA(RML) [19] &66.00\% & 55\\
GCCA(RML) [14] &73.96\% & 26\\
CCA(RML) [11] &60.42\% & 48\\
\textbf{LMCCA(eNT)} & \textbf{93.75\%} & 28\\
HDLM(eNT) [55] &85.97\% &---\\
MCCA(eNT) [19] &82.29\% & 25\\
GCCA(eNT) [14] &82.29\% & 19\\
CCA(eNT) [11] &71.88\% & 20\\
\hline
\end{tabular}
\end{table}
\subsection{Performance of Object Recognition}
Since MNIST, ORL, RML and eNT are all fairly small or medium datasets, we further validated our method on Caltech 101 dataset which consists of images from 101 object categories, containing from 31 to 800 images per category. The size of each image is roughly 300 $\times$ 200 pixels. The significant variation in color, pose and lighting makes this dataset challenging. A number of previously published papers have reported results on this data set [44-48]. Some sample images of nine categories from the Caltech 101 dataset are given in Fig. 9.
\centerline {\includegraphics[width=2.2in]{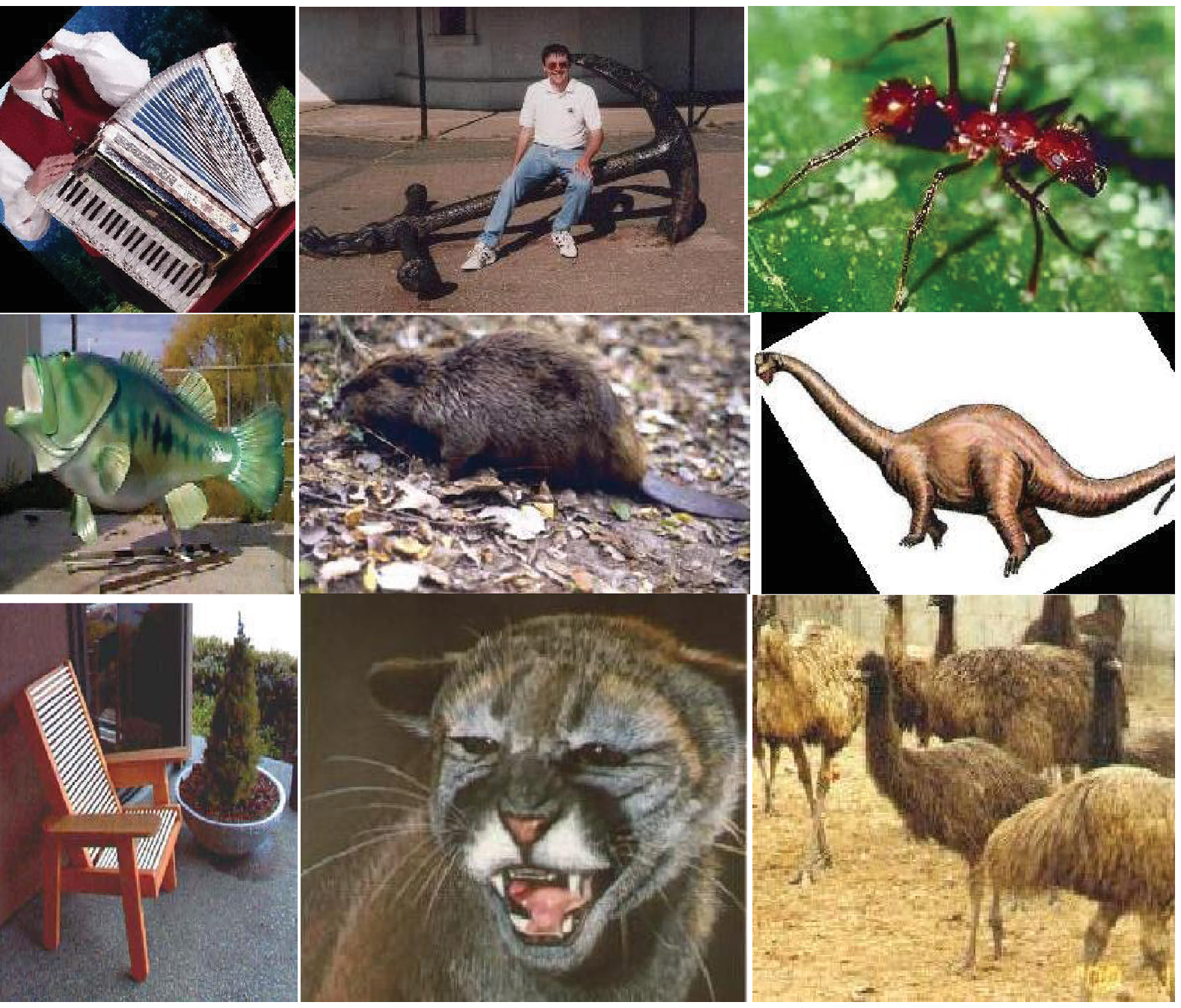}}\\ \centerline {Fig. 9 Images from the Caltech 101 database}\\\indent

For fair comparisons, we performed experiments as in other studies [44-48]. Specifically, for each trial, we selected 30 images from each class as training samples. The remaining images from each class were tested, and the average recognition accuracy was reported. The performance of GIST [44], PHOG [45], and LBP [46] is tabulated in TABLE IX.
\vspace*{-10pt}
\begin{table}[h]
\small
\renewcommand{\arraystretch}{1.4}
\caption{\normalsize{Results of Object Recognition with A Single Feature (Caltech 101)}}
\setlength{\abovecaptionskip}{0pt}
\setlength{\belowcaptionskip}{10pt}
\centering
\tabcolsep 0.073in
\begin{tabular}{cc}
\hline
Single Feature & Recognition Accuracy\\
\hline
GIST [44] &20.97\%\\
PHOG [45] &20.15\%\\
LBP [46] &20.77\%\\
\hline
\end{tabular}
\end{table}\\

TABLE IX suggests we should use GIST (\textbf{20.97\%}) and LBP (\textbf{20.77\%}) which provide the best individual performance as the input to CCA and GCCA. 
The performance by the methods of CCA, GCCA, MCCA, and LMCCA is shown in Fig. 10. Note, LMCCA achieves the optimal performance when the projected dimension \emph{d} is equal to 48, which is less than Q=20+59+40=119. DL based methods [42-43] and DL+SML methods [49-50] are also investigated on the Caltech 101 dataset. The parameters of the DL based approaches are described as follows:\\
\textbf{CNN}: iterations=5000 and learning rate is 0.005.\\
\textbf{PCANet}: the filter size $k_1$=$k_2$=7 and the number of filters in each stage $L_1$=$L_2$=8.\\
\textbf{DEFEATnet}: the dictionary size is fixed to be 1024, 2048, and 4096 in the layer 1, layer 2, and layer 3, respectively.\\
\textbf{SP Pooling}: the number level of pyramid structure is four and the pyramid is {6 $\times$ 6, 3 $\times$ 3, 2 $\times$ 2, 1 $\times$ 1} (totally 50 bins). In addition, the standard 10-view prediction with each view a 224 $\times$ 224 crop is used in our experiments.\\\indent
The relation between highest recognition accuracy and dimension with different methods is tabulated in TABLE X which brings out some interesting observations:\\
\textbf{1.} Even on this challenging dataset, the performance of most of the pure SML methods (MCCA, EP, Visual Cortex and LMCCA) and that of the pure DL methods (CNN and PCANet) are essentially the same, with LMCCA being the winner.\\
\textbf{2.} The DL+SML methods, DEFEATnet and SP Pooling, are hybrid of SML and DL, with DEFEATnet incorporating prior knowledge into DL and SP Pooling combining spatial pyramid pooling [56-57] and CNN method.\\\indent
\centerline {\includegraphics[width=3.8in]{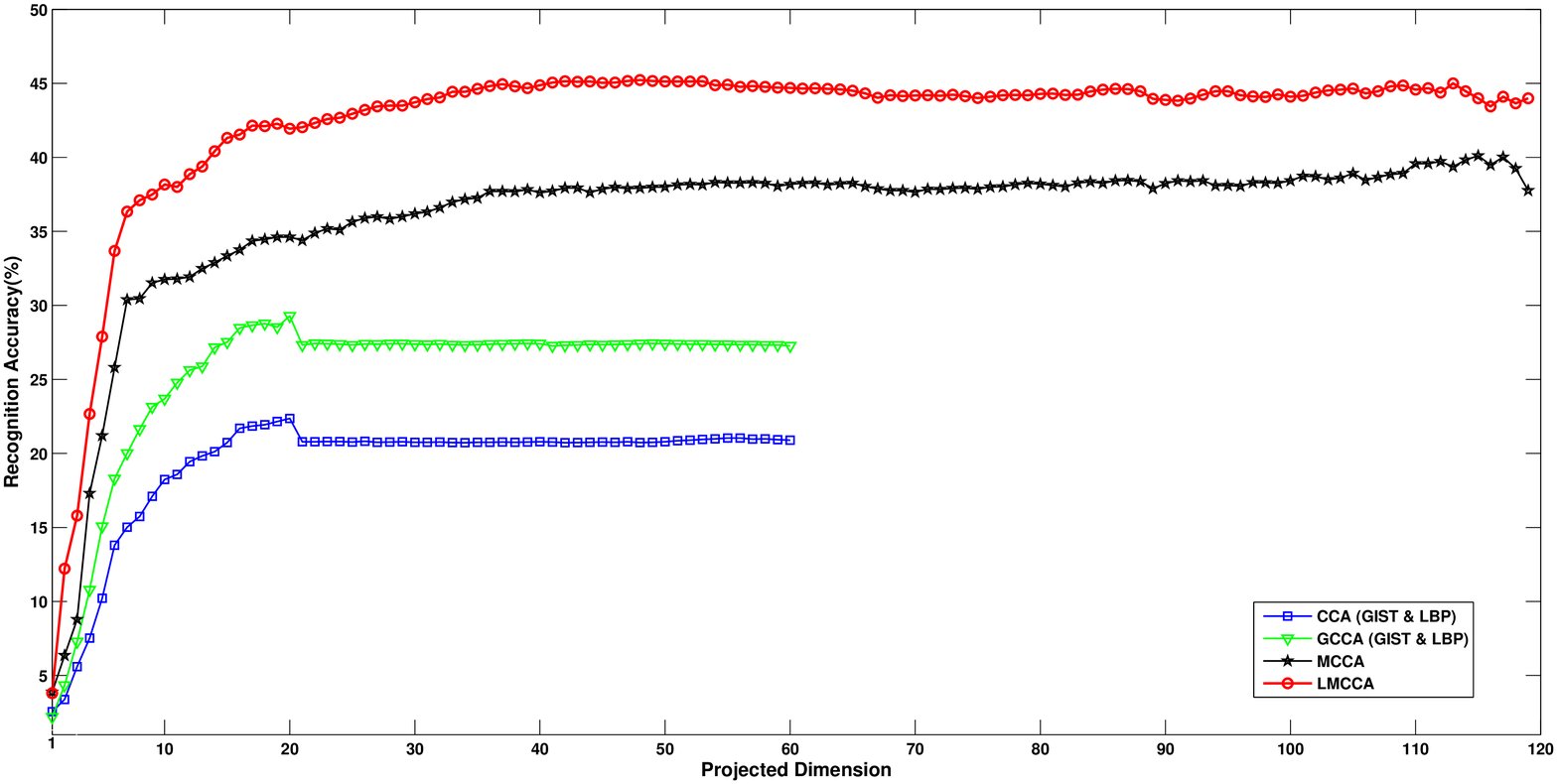}}\\ \leftline {Fig. 10 Experimental Results on The Caltech 101 Database}\\\indent
\vspace*{-10pt}
\begin{table}[h]
\small
\renewcommand{\arraystretch}{1.4}
\caption{\normalsize{Relation between optimal accuracy and dimension (Caltech 101)}}
\setlength{\abovecaptionskip}{0pt}
\setlength{\belowcaptionskip}{10pt}
\centering
\tabcolsep 0.07in
\begin{tabular}{ccc}
\hline
Method & Highest Accuracy & Dimension(Number)\\
\hline
Serial Fusion [9] &27.34\% & ---\\
CCA [11] &22.37\% & 20\\
GCCA [14] &29.29\% & 20\\
MCCA [19] &40.11\% & 115\\
CNN [42] &38.45\% & ---\\
PCANet [43] &42.99\% & ---\\
EP Method [47] &33.00\% & ---\\
Visual Cortex [48] &42.00\% & ---\\
DEFEATnet [49] &77.60\% & ---\\
SP Pooling [50] &91.44\% & ---\\
\textbf{LMCCA} & \textbf{45.23\%} & 48\\
\hline
\end{tabular}
\end{table}

To further demonstrate the effectiveness of the proposed method, we extracted three different features from DL networks and LMCCA is applied to develop new representation from the extracted DL features. Alex network is applied to our experiments and the extracted features of three fully connected layers \textbf{fc6}, \textbf{fc7} and \textbf{fc8} are used for fusion. The performance is given in TABLE XI.\\
\vspace*{-10pt}
\begin{table}[h]
\small
\renewcommand{\arraystretch}{1.4}
\caption{\normalsize{The performance of different features from \textnormal{fc6}, \textnormal{fc7} and \textnormal{fc8} (Caltech 101)}}
\setlength{\abovecaptionskip}{0pt}
\setlength{\belowcaptionskip}{10pt}
\centering
\tabcolsep 0.07in
\begin{tabular}{ccc}
\hline
Feature & Performance\\
\hline
\textbf{fc6} & 77.84\% \\
\textbf{fc7} & 77.65\% \\
\textbf{fc8} & 73.31\% \\
\hline
\end{tabular}
\end{table}

Then, the proposed LMCCA is applied to the three extracted features before recognition is performed. The comparison with other related methods is shown in TABLE XII.\\
\vspace*{-10pt}
\begin{table}[h]
\small
\renewcommand{\arraystretch}{1.4}
\caption{\normalsize{Comparison with other methods (Caltech 101)}}
\setlength{\abovecaptionskip}{0pt}
\setlength{\belowcaptionskip}{10pt}
\centering
\tabcolsep 0.07in
\begin{tabular}{ccc}
\hline
Methods & Performance & Method Types\\
\hline
B. Du et al. (2017) [59] &78.60\% & DL fusion\\
L. Mansourian et al. (2017) [60] &75.37\% & SML fusion\\
P. Tang et al. (2017) [61] &82.45\% & SML+DL fusion\\
G. Lin et al. (2017) [62] &78.73\% & SML fusion\\
W. Xiong et al. (2017) [63] &75.90\% & SML fusion\\
S. Kim et al. (2017) [64] &83.00\% & SML+DL fusion\\
W. Yu et al. (2018) [65] &77.90\% & SML+DL fusion\\
L. Sheng et al. (2018) [66] &74.78\% & SML fusion\\
C. Ma et al. (2018) [67] &43.52\% & SML fusion\\
\textbf{Our method (LMCCA)} & \textbf{83.68\%} & SML+DL fusion\\
\hline
\end{tabular}
\end{table}

In TABLE XII, we compared LMCCA with nine methods belonging to three classes: 1) Features extracted by DL and fused by neural nets (DL fusion); 2) Feature extracted by classical methods and fused by SML (SML fusion); 3) Features extracted by both classical and DL and fused by SML (SML+DL fusion). With LMCCA in fusion, the proposed method outperforms all the others, demonstrating the effectiveness of the proposed method in information fusion and its suitability to work with different feature extraction methods. Note the other good performers [61, 64] are also based on SML+DL fusion principles. The comparison clearly demonstrates the proposed LMCCA is capable of improving the quality of features extracted by DL, leading to enhanced recognition performance.
\section{CONCLUSIONS}
In this paper, we propose a generic method for information fusion based on the labeled multiple canonical correlations analysis (LMCCA) and establish a unified framework for canonical correlation analysis with intra-class information. Furthermore, we analytically and experimentally verify the upper limit on the number of projected dimensions of LMCCA to achieve the best recognition performance. Since the dimension of the best performance has no relation to the number of training samples, it can form a practical platform to deal with the big data fusion problems. To demonstrate the effectiveness and generic nature of LMCCA, the method is used to different applications. Experimental results demonstrate that the proposed method outperforms the related SML methods and some DL methods on all small/medium datasets. \\\indent We further studied fusion based on LMCCA and DL principles, and applied to object recognition on the Caltech 101 dataset. The results clearly demonstrate the effectiveness of methodology fusion.

\appendices
\section{}
The solutions to matrix transformation (e.g., LDA and PCA) are usually the eigenvectors associated with the eigenvalues in a form similar to that of equation (22):
\begin{equation}
\frac{1}{{P - 1}}inv(F) \times (G-F)\omega  = \lambda \omega
\end{equation}
where \textit{inv()} refers to the inverse transform of a matrix. However, unless the matrix \textit{F} has full rank, the block matrix in equation (38) is singular. A popular approach [40] of dealing with singular matrices and controlling complexity is to add a multiple of the identity matrix $\rho {\rm I}(\rho  > 0)$ to \textit{F}. Thus, the generalized form of equation (38) is written as:
\begin{equation}
\frac{1}{{P - 1}}inv({F^ + }) \times (G - {F})\omega  = \eta \omega
\end{equation}
where
\begin{equation}
\begin{array}{*{20}{c}}
   {{F^ + } = \left\{ {\begin{array}{*{20}{c}}
   {F\;   {\rm{when}}\; F\; {\rm{is}}\; {\rm{a}}\; {\rm{non-singular}}\; {\rm{matrix}}}  \\
   \begin{array}{l}
 F + \rho {\rm{I }}\quad {\rm{when}}\; F\; {\rm{is}}\; {\rm{a}}\; {\rm{singular}}\; {\rm{matrix}} \\
 {\rm{I}} \in {R^{Q \times Q}} \\
 \end{array}  \\
\end{array}} \right.}  \\

\end{array}
\end{equation}\\\indent
Equation (39) is further written as:
\begin{small}
\begin{equation}
\begin{array}{*{20}{c}}
   {F_1^ + {R_{{x_1}{x_2}}}{\omega _{{x_2}}} + F_1^ + {R_{{x_1}{x_3}}}{\omega _{{x_3}}} +  \cdots  + F_1^ + {R_{{x_1}{x_P}}}{\omega _{{x_P}}} = {\eta}{\omega _{{x_1}}}}  \\
   {F_2^ + {R_{{x_2}{x_1}}}{\omega _{{x_1}}} + F_2^ + {R_{{x_2}{x_3}}}{\omega _{{x_3}}} +  \cdots  + F_2^ + {R_{{x_2}{x_P}}}{\omega _{{x_P}}} = {\eta}{\omega _{{x_2}}}}  \\
    \vdots   \\
   {F_P^ + {R_{{x_P}{x_1}}}{\omega _{{x_1}}} + F_P^ + {R_{{x_P}{x_2}}}{\omega _{{x_2}}} +  \cdots  + F_P^ + {R_{{x_P}{x_{P - 1}}}}{\omega _{{x_{P - 1}}}} = {\eta}{\omega _{{x_P}}}}  \\
\end{array}
\end{equation}
\end{small}\\\indent
Rearranging equation (41) leads to
\begin{equation}
\begin{array}{*{20}{c}}
   {F_1^ + ({R_{{x_1}{x_2}}}{\omega _{{x_2}}} + {R_{{x_1}{x_3}}}{\omega _{{x_3}}} +  \cdots  + {R_{{x_1}{x_P}}}{\omega _{{x_P}}}) = {\eta}{\omega _{{x_1}}}}  \\
   {F_2^ + ({R_{{x_2}{x_1}}}{\omega _{{x_1}}} + {R_{{x_2}{x_3}}}{\omega _{{x_3}}} +  \cdots  + {R_{{x_2}{x_P}}}{\omega _{{x_P}}}) = {\eta}{\omega _{{x_2}}}}  \\
    \vdots   \\
   {F_P^ + ({R_{{x_P}{x_1}}}{\omega _{{x_1}}} + {R_{{x_P}{x_2}}}{\omega _{{x_2}}} +  \cdots  + {R_{{x_P}{x_{P - 1}}}}{\omega _{{x_{P - 1}}}}) = {\eta}{\omega _{{x_P}}}}  \\
\end{array},
\end{equation}
where
\begin{small}
\begin{equation}
{F_q}^ +  = \left\{ \begin{array}{l}
 S{\omega _{{x_q}}}\qquad \quad {\rm{when}}\; S{\omega _{{x_q}}}\; {\rm{is}}\; {\rm{a}}\; {\rm{non-singular}}\; {\rm{matrix}} \\
 S{\omega _{{x_q}}} + \rho {I_q}\  {\rm{when}}\; S{\omega _{{x_q}}}\; {\rm{is}}\; {\rm{a}}\; {\rm{singular}}\; {\rm{matrix}} \\
 ({I_q} \in {R^{\:{m_q} \times {m_q}}}, q = 1,2...P) \\
 \end{array} \right.,
\end{equation}
\end{small} and ${S{\omega_{{{x_q}}}}} $ denotes the within-class scatter matrix of $ {{x_q}} $ in the form of $R^{{m_q}\times {m_q}}$. ${S{\omega_{{{x_q}}}}} $ is explicitly given in the following form\\
\begin{equation}\ S{\omega_{{x_q}}} = \sum\limits_{i = 1}^c {p({\omega _{i}})\left[ {\sum\limits_{j = 1}^{{l_i}} {\frac{1}{{{l_i}}}({x_q}_{ij} - {m^{{x_q}}}_i){{({x_q}_{ij} - {m^{{x_q}}}_i)}^T}} } \right]} \end{equation}\\\indent From equation (42), the rank of $\omega _{{x_q}} (q=1,2,...P)$ satisfies
\begin{equation} \ rank(\omega _{{x_q}}) \le rank({F_q}^ +) \end{equation}\\
Now, we analyze the following two cases.\\

\textbf{1) when ${S{\omega_{{{x_q}}}}} $ is a full rank matrix}\\\indent
Since $ p({\omega _i})$ and ${l_i}$ are constants without any influence on the rank of within-class scatter matrix, equation (44) is further expanded to \\\
\begin{equation} \ S{w_{x_q}} = {S_{x_q}}{S_{x_q}}^T \end{equation}
where ${S_{x_q}}$ is in the following form
\begin{equation}\ S{_{{x_q}}} = \left[ {{S^1}_{{x_q}}, {S^2}_{{x_q}}, ..., {S^c}_{{x_q}}} \right]
\end{equation}
\begin{equation} \begin{array}{*{20}{c}}
   {{S^1}_{{x_q}} = \left[ {{x_q}_{(11)} - {m^{{x_q}}}_1,{x_{q}}_{(12)} - {m^{{x_q}}}_1,...{x_{q}}_{(1{l_1})} - {m^{{x_q}}}_1} \right]}  \\
   {{S^2}_{{x_q}} = \left[ {{x_q}_{(21)} - {m^{{x_q}}}_2,{x_{q}}_{(22)} - {m^{{x_q}}}_2,...{x_q}_{(2{l_2})} - {m^{{x_q}}}_2} \right]}  \\
    \vdots   \\
   {{S^c}_{{x_q}} = \left[ {{x_{q}}_{(c1)} - {m^{{x_q}}}_c,{x_q}_{(c2)} - {m^{{x_q}}}_c,...{x_{q}}_{(c{l_c})} - {m^{{x_q}}}_c} \right]}  \\
\end{array}
 \end{equation}
\\\indent Adding all the column vectors of $ {S^1}_{{x_q}}\ $, $ {S^2}_{{x_q}}\ $,...$ {S^c}_{{x_q}}\ $ to the first column yields
\begin{equation} \begin{array}{*{20}{c}}
   {{S^1}_{{x_q}} = \left[ {0,{x_{q}}_{(12)} - {m^{{x_q}}}_1,...{x_{q}}_{(1{l_1})} - {m^{{x_q}}}_1} \right]}  \\
   {{S^2}_{{x_q}} = \left[ {0,{x_{q}}_{(22)} - {m^{{x_q}}}_2,...{x_q}_{(2{l_2})} - {m^{{x_q}}}_2} \right]}  \\
    \vdots   \\
   {{S^c}_{{x_q}} = \left[ {0,{x_q}_{(c2)} - {m^{{x_q}}}_c,...{x_{q}}_{(c{l_c})} - {m^{{x_q}}}_c} \right]}  \\
\end{array}
 \end{equation}
\\\indent
\begin{equation}
rank({S_{{x_q}}}) \le N - c
\end{equation}
where $N$ is the number of the total training samples and $c$ is the number of classes.\\
\\\indent Therefore
\begin{equation}
rank(S{\omega _{{x_q}}}) \le \min [rank({S_{{x_q}}}),rank({S^T}_{{x_q}})] \le N - c
\end{equation}
\\\indent Besides, since ${S{\omega_{{{x_q}}}}} $ is in the form of $R^{{m_q}\times {m_q}}$, the rank of $S{\omega _{{x_q}}}$ satisfies the following relation
\begin{equation}
rank(S{\omega _{{x_q}}}) \le {m_q}
\end{equation}
\\\indent Since ${S{\omega_{{{x_q}}}}} $ is a full rank matrix, the dimension of $x_q$ satisfies inequality (53)
\begin{equation}
{m_q}\le N-c,
\end{equation}
combining (51) and (52) shows that the rank of ${{F_q}^ +}$ satisfies the following relation
\begin{equation}
rank({{F_q}^ +}) \le {m_q}
\end{equation}\\\indent
\textbf{2) when ${S{\omega_{{{x_q}}}}}$ is a singular matrix}
\begin{equation}
{F_q}^ +  = S{\omega _{{x_q}}} + \rho {I_q}({I_q} \in {R^{{m_q} \times {m_q}}}, q = 1,2...P)
\end{equation}
obviously,
\begin{equation}
rank({{F_q}^ +}) \le {m_q}
\end{equation}
\\\indent Thus, in summary, the rank of ${{F_q}^ +}$ satisfies $rank({{F_q}^ +}) \le {m_q}$ under each of the two different conditions.\\\indent Therefore,
\begin{equation}
\ rank(\omega _{{x_q}}) \le rank({{F_q}^ +}) \le {m_q}
\end{equation}

Since there are \emph{P} sets of features, the rank of $\omega$ satisfies
\begin{equation}
{{rank({\omega}}) }{\rm{ }} \le {rank({{F_1}^ +}})+{rank({{F_2}^ +}})+...+{rank({{F_P}^ +}})
\end{equation}\\\indent
Thus
\begin{equation}
{{rank({\omega}}) }{\rm{ }} \le {m_1}+{m_2}+...+{m_P}=Q
\end{equation}
\\\indent Therefore, the number of projected dimension \emph{d} corresponding to the optimal recognition accuracy satisfies:
\begin{equation}
d \le Q
\end{equation}

\section*{Acknowledgment}

This work is supported by the National Natural Science Foundation of China (NSFC, No.61071211), the State Key Program of NSFC (No. 61331021), the Key International Collaboration Program of NSFC (No. 61210005) and the Discovery Grant of Natural Science and Engineering Council of Canada (No. 238813/2010).

\ifCLASSOPTIONcaptionsoff
  \newpage
\fi


\begin{thebibliography}{1}
\bibitem{IEEEhowto:kopka}
X. Chen et al. ``Multimodal video indexing and retrieval using directed information." \emph{IEEE TMM}, vol. 14, no. 1, pp. 3--16, 2012.
\bibitem{IEEEhowto:kopka}
B. Khaleghi et al. ``Multisensor data fusion: A review of the state-of-the-art." \emph{Information Fusion}, vol. 14, no.1, pp. 28--44, 2013.
\bibitem{IEEEhowto:kopka}
Y. Yang et al. ``Multi-feature fusion via hierarchical regression for multimedia analysis." \emph{IEEE TMM}, vol. 15, no. 3, pp. 572--581, 2013.
\bibitem{IEEEhowto:kopka}
G. Evangelopoulos et al. ``Multimodal saliency and fusion for movie summarization based on aural, visual, and textual attention." \emph{IEEE TMM}, vol. 15, no. 7, pp. 1553--1568, 2013.
\bibitem{IEEEhowto:kopka}
C.S. Wu et al. ``Two-level hierarchical alignment for semi-coupled HMM-based audiovisual emotion recognition with temporal course." \emph{IEEE TMM}, vol. 15, no. 8, pp. 1880--1895, 2013.
\bibitem{IEEEhowto:kopka}
Y. Yeh et al. ``A novel multiple kernel learning framework for heterogeneous feature fusion and variable selection." \emph{IEEE TMM}, vol. 14, no. 3, pp. 563--574, 2012.
\bibitem{IEEEhowto:kopka}
L. Nanni et al. ``Combining biometric matchers by means of machine learning and statistical approaches." \emph{Neurocomputing}, vol. 149, pp. 526--535, 2015.
\bibitem{IEEEhowto:kopka}
C. Ding et al. ``Robust face recognition via multimodal deep face representation." \emph{IEEE TMM}, vol. 17, no. 11, pp. 2049--2058, 2015.
\bibitem{IEEEhowto:kopka}
J. Yang et al. ``Feature-level fusion of fingerprint and finger-vein for personal identification." \emph{Pattern Recognition Letters}, vol. 33, no. 5, pp. 623--628, 2012.
\bibitem{IEEEhowto:kopka}
J. Yang et al. ``Feature fusion: parallel strategy vs. serial strategy."  \emph{Pattern Recognition}, vol. 36, no. 6, pp. 1369--1381, 2003.
\bibitem{IEEEhowto:kopka}
L. Sun et al. ``Canonical correlation analysis for multilabel classification: a least-squares formulation, extensions, and analysis." \emph{IEEE TPAMI}, vol. 33, no. 1, pp. 194--200, 2011.
\bibitem{IEEEhowto:kopka}
J. Ver et al. ``Deterministic CCA-based algorithms for blind equalization of FIR-MIMO channels." \emph{IEEE TSP}, vol. 55, no. 7, pp. 3867--3878, 2007.
\bibitem{IEEEhowto:kopka}
H. Izadinia et al. ``Multimodal analysis for identification and segmentation of moving-sounding objects." \emph{IEEE TMM}, vol. 15, no. 2, pp. 378--390, 2013.
\bibitem{IEEEhowto:kopka}
Q. Sun et al. ``A theorem on the generalized canonical projective vectors." \emph{Pattern Recognition}, vol. 38, no. 10, pp. 449--452, 2005.
\bibitem{IEEEhowto:kopka}
 A. Tenenhaus et al. ``Regularized generalized canonical correlation analysis for multiblock or multigroup data analysis." \emph{European Journal of operational research}, vol. 238, no. 2, pp. 391--403, 2014.
\bibitem{IEEEhowto:kopka}
C. Shen et al. ``Generalized canonical correlation analysis for classification." \emph{Journal of Multivariate Analysis}, vol. 130, pp. 310--322, 2014.
\bibitem{IEEEhowto:kopka}
X. Yang et al. ``Cross-domain feature learning in multimedia." \emph{IEEE TMM}, vol. 17, no. 1, pp. 64--78, 2015.
\bibitem{IEEEhowto:kopka}
Y.O. Li et al. ``Joint blind source separation by multiset canonical correlation analysis." \emph{IEEE TSP}, vol. 57, no. 10, pp. 3918--3929, 2009.
\bibitem{IEEEhowto:kopka}
H.G. Yu et al. ``Multiset Canonical Correlation Analysis Using for Blind Source Separation." \emph{Applied Mechanics and Materials}, vol. 195, pp. 104--108, 2012.
\bibitem{IEEEhowto:kopka}
Y.O. Li et al. ``Group study of simulated driving fMRI data by multiset canonical correlation analysis." \emph{Journal of signal processing systems}, vol. 68, no. 1, pp. 31--48, 2012.
\bibitem{IEEEhowto:kopka}
P. N. Belhumeur et al. ``Eigenfaces vs. Fisherfaces: Recognition Using Class Specific Linear Projection." \emph{IEEE TPAMI}, vol. 19, no. 7, pp. 711--720, Jul. 1997.
\bibitem{IEEEhowto:kopka}
A. M. Mart¨ªnez et al. ``Pca versus lda." \emph{IEEE TPAMI}, vol. 23, no. 2, pp. 228-233, 2001.
\bibitem{IEEEhowto:kopka}
A. Khotanzad et al. ``Invariant image recognition by Zernike moments." \emph{IEEE TPAMI}, no. 12, pp. 489--497, 1990.
\bibitem{IEEEhowto:kopka}
Y. Hamamoto et al. ``A gabor filter-based method for recognizing handwritten numerals." \emph{Pattern Recognition}, vol. 31, no. 4, pp. 395--400, 1998.
\bibitem{IEEEhowto:kopka}
A. Satpathy et al. ``LBP-based edge-texture features for object recognition." \emph{IEEE TIP}, vol. 23, no. 5, pp. 1953--1964, 2014.
\bibitem{IEEEhowto:kopka}
D. Huang et al. ``HSOG: a novel local image descriptor based on histograms of the second-order gradients." \emph{IEEE TIP}, vol. 23, no. 11, pp.4680-4695, 2014.
\bibitem{IEEEhowto:kopka}
M. Yang et al. ``Gabor feature based sparse representation for face recognition with gabor occlusion dictionary." \emph{ECCV}, pp. 448--461, 2010.
\bibitem{IEEEhowto:kopka}
M. El Ayadi et al. ``Survey on speech emotion recognition: Features, classification schemes, and databases." \emph{Pattern Recognition}, vol. 44, no. 3, pp. 572--587, 2011.
\bibitem{IEEEhowto:kopka}
C.H. Wu et al. ``Emotion recognition of affective speech based on multiple classifiers using acoustic-prosodic information and semantic labels." \emph{IEEE TAC}, vol. 2, no. 1, pp. 10--21, 2011.
\bibitem{IEEEhowto:kopka}
C.S. Ooi et al. ``A new approach of audio emotion recognition." \emph{Expert Systems with Applications}, vol. 41, no. 13, pp. 5858--5869, 2014.
\bibitem{IEEEhowto:kopka}
Y. Zhou et al.``Deception detecting from speech signal using relevance vector machine and non-linear dynamics features." \emph{Neurocomputing} vol. 151, pp. 1042--1052, 2015.
\bibitem{IEEEhowto:kopka}
R. Cowie et al. ``Emotion recognition in human-computer interaction." \emph{IEEE SPM}, vol. 18, no. 1, pp. 32--80, 2001.
\bibitem{IEEEhowto:kopka}
Y. Wang et al. "Recognizing human emotion from audiovisual signals." \emph{IEEE TMM}, vol. 10, no. 5, pp. 936--946, August 2008.
\bibitem{IEEEhowto:kopka}
M. J. Black et al. ``Recognizing facial expressions in image sequences using local parameterized models of image motion." \emph{IJCV}, vol. 25, no. 1, pp. 23--48, 1997.
\bibitem{IEEEhowto:kopka}
S. M. Lajevardi et al. ``Facial expression recognition in perceptual color space." \emph{IEEE TIP}, vol. 21, no. 8, pp. 3721--3733, 2012.
\bibitem{IEEEhowto:kopka}
Y. Tie et al. ``Human emotional state recognition using real 3D visual features from Gabor library." \emph{Pattern Recognition}, vol. 46, pp. 529--538, 2013.
\bibitem{IEEEhowto:kopka}
R. Xiao et al. ``Facial expression recognition on multiple manifolds." \emph{Pattern Recognition}, vol. 44, no. 1, pp. 107--116, 2011.
\bibitem{IEEEhowto:kopka}
B. H. Shekar et al. ``Face recognition using kernel entropy component analysis." \emph{Neurocomputing}, vol. 74, no. 6, pp. 1053--1057, 2011.
\bibitem{IEEEhowto:kopka}
O. Martin et al. ``The enterface audio-visual emotion database." \emph{22nd International Conference on Data Engineering Workshops (ICDEW)}, IEEE, pp. 1--8, 2006.
\bibitem{IEEEhowto:kopka}
T. Melzer et al. ``Appearance models based on kernel canonical correlation analysis." \emph{Pattern recognition}, vol. 36, no. 9 pp: 1961--1971, 2003.
\bibitem{IEEEhowto:kopka}
Y. Wang et al. ``Kernel based fusion with application to audiovisual emotion recognition." \emph{IEEE TMM}, vol. 14, no. 3, pp. 597--607, 2012.
\bibitem{IEEEhowto:kopka}
A. Krizhevsky et al. ``ImageNet classification with deep convolutional neural network."\emph{In Advances in neural information processing systems}, pp. 1097--1105, 2012.
\bibitem{IEEEhowto:kopka}
T.H. Chan et al. ``PCANet: A simple deep learning baseline for image classification?" \emph{IEEE TIP}, vol. 24, no. 12, pp. 5017--5032, 2015.
\bibitem{IEEEhowto:kopka}
A. Oliva et al. ``Modeling the shape of the scene: A holistic representation of the spatial envelope." \emph{IJCV}, vol. 42, no. 3, pp. 145--175, 2001.
\bibitem{IEEEhowto:kopka}
A. Bosch et al. ``Image classification using random forests and ferns." \emph{IEEE 11th ICCV}, pp. 1--8, 2007.
\bibitem{IEEEhowto:kopka}
T. Ojala et al. ``Multiresolution gray-scale and rotation invariant texture classification with local binary patterns." \emph{IEEE TPAMI}, vol. 24, no. 7, pp. 971--987, 2002.
\bibitem{IEEEhowto:kopka}
D. Dai et al. ``Ensemble projection for semi-supervised image classification." \emph{In Proceedings of the IEEE ICCV}, pp. 2072--2079, 2013.
\bibitem{IEEEhowto:kopka}
T. Serre et al. ``Object recognition with features inspired by visual cortex." \emph{IEEE CVPR}, vol. 2, pp. 994--1000, 2005.
\bibitem{IEEEhowto:kopka}
S. Gao et al. ``DEFEATnet deep conventional image representation for image classification." \emph{IEEE TCSVT}, vol. 26, no. 3, pp. 494--505, 2016.
\bibitem{IEEEhowto:kopka}
K. He et al. ``Spatial pyramid pooling in deep convolutional networks for visual recognition." \emph{IEEE TPAMI}, vol. 37, no. 9, pp. 1904--1916, 2015.
\bibitem{IEEEhowto:kopka}
Y. Xu et al. ``A New Discriminative Sparse Representation Method for Robust Face Recognition via $l_1$ Regularization." \emph{IEEE TNNLS}, vol. 28, no. 10, pp. 2233--2241, 2017.
\bibitem{IEEEhowto:kopka}
L. Zhang et al. ``Sparse representation or collaborative representation: Which helps face recognition?" \emph{2011 IEEE ICCV}, pp. 471--478, 2011.
\bibitem{IEEEhowto:kopka}
$l1_ls$: Simple MATLAB Solver for l1-regularized Least Squares Problems. [Online]. Available:
http://web.stanford.edu/~boyd/$l1_ls$/.
\bibitem{IEEEhowto:kopka}
A.Y. Yang et al. ``Fast $\ell_ {1} $-Minimization Algorithms for Robust Face Recognition." \emph{IEEE TIP}, vol. 22, no. 8, pp. 3234--3246, 2013.
\bibitem{IEEEhowto:kopka}
S. Zhang et al. ``Learning Affective Features with a Hybrid Deep Model for Audio-Visual Emotion Recognition." \emph{IEEE TCSVT (Accept)},  2017.
\bibitem{IEEEhowto:kopka}
K. Grauman et al. ``The pyramid match kernel: Discriminative classification with sets of image features." \emph{2005 IEEE ICCV}, vol. 2, pp. 1458--1465, 2005.
\bibitem{IEEEhowto:kopka}
S. Lazebnik et al. ``Beyond bags of features: Spatial pyramid matching for recognizing natural scene categories." \emph{2006 IEEE CVPR}, vol. 2, pp. 2169--2178, 2006.
\bibitem{IEEEhowto:kopka}
L. Gao et al. ``Discriminative Multiple Canonical Correlation Analysis for Information Fusion." \emph{IEEE TIP}, vol. 27. no. 4, pp. 1951--1965, 2017.
\bibitem{IEEEhowto:kopka}
B. Du et al. ``Stacked convolutional denoising auto-encoders for feature representation." \emph{IEEE Transactions on cybernetics}, vol. 47, no. 4, pp. 1017-1027, 2017.
\bibitem{IEEEhowto:kopka}
 L. Mansourian et al. ``An effective fusion model for image retrieval." \emph{Multimedia Tools and Applications}, pp. 1-24, 2017.
\bibitem{IEEEhowto:kopka}
 P. Tang et al. ``Learning multi-instance deep discriminative patterns for image classification." \emph{IEEE TIP}, vol. 26, no. 7, pp. 3385-3396, 2017.
 \bibitem{IEEEhowto:kopka}
 G. Lin et al. ``Visual feature coding based on heterogeneous structure fusion for image classification." \emph{Information Fusion}, vol. 36, pp. 275-283, 2017.
 \bibitem{IEEEhowto:kopka}
 W. Xiong et al. ``Combining local and global: Rich and robust feature pooling for visual recognition." \emph{Pattern Recognition}, vol. 62, pp. 225-235, 2017.
 \bibitem{IEEEhowto:kopka}
 S. Kim et al. ``Fcss: Fully convolutional self-similarity for dense semantic correspondence." \emph{IEEE CVPR}, vol. 1, no. 2, pp.1--8, 2017.
 \bibitem{IEEEhowto:kopka}
 W. Yu et al. "Hierarchical semantic image matching using CNN feature pyramid." \emph{Computer Vision and Image Understanding}, vol. 167, pp. 40--51, 2018.
 \bibitem{IEEEhowto:kopka}
 S. Li et al. "Self-taught low-rank coding for visual learning." \emph{IEEE TNNLS}, vol. 29, no. 3, pp. 645--656, 2018.
 \bibitem{IEEEhowto:kopka}
C. Ma et al. "Error Correcting Input and Output Hashing." \emph{IEEE Transactions on Cybernetics (Accept)}, 2018.










\bibitem{IEEEhowto:kopka}
 K. Kumar et al. "F-DES: Fast and Deep Event Summarization." \emph{IEEE TMM} vol. 20, no. 2, pp. 323-334, 2018.
\bibitem{IEEEhowto:kopka}
N. Singh et al. "A convex hull approach in conjunction with Gaussian mixture model for salient object detection." \emph{Digital Signal Processing}, vol. 55, pp. 22-31, 2016.



\end{thebibliography}
\end{document}